\pdfoutput=1

\documentclass[11pt]{article}

\usepackage{emnlp2021}

\usepackage{times}
\usepackage{latexsym}
\usepackage{booktabs}
\usepackage{amsmath}

\usepackage[T1]{fontenc}

\usepackage[utf8]{inputenc}

\usepackage{microtype}

\usepackage{xspace}
\usepackage{booktabs}
\usepackage{graphicx}
\usepackage{enumitem}
\usepackage{xcolor,soul}
\usepackage{tikz-dependency}

\usepackage{url}
\usepackage{amssymb}
\usepackage{amsmath}
\usepackage{float}
\usepackage{todonotes}
\usepackage[dash,dot]{dashundergaps}

\usepackage[normalem]{ulem}
\makeatletter
\newcommand{\rAB}[2]{\bgroup \UL@setULdepth
 \markoverwith{\lower\ULdepth\hbox
   {\kern-.02em\vbox{\color{#1}\hrule width.2em\kern1.2\p@\color{#2}\hrule}\kern-.02em}}%
 \ULon}
\makeatother
\usepackage{amsmath}
\usepackage{pifont}
\newcommand{\cmark}{\text{\ding{51}}}
\newcommand{\xmark}{\text{\ding{55}}}

\definecolor{darkgreen}{RGB}{76,153,0}

\sethlcolor{lightgray}

\usepackage{multirow}
\usepackage{array}
\newcolumntype{C}[1]{>{\centering\let\newline\\\arraybackslash\hspace{0pt}}m{#1}}

\setlist[description]{leftmargin=\parindent,labelindent=0pt}

\newcommand{\red}[1]{\textcolor{red}{#1}}

\newcommand{\enquote}[1]{``#1''}

\newcommand{\fs}{F\textsubscript{1}\xspace}
\newcommand{\nispar}{NIS\textsubscript{tok}\xspace}

\newcommand{\sref}[1]{Sec.~\ref{sec:#1}}
\newcommand{\aref}[1]{Appendix~\ref{sec:#1}}

\newcommand{\tref}[1]{Table~\ref{tab:#1}}
\newcommand{\fref}[1]{Figure~\ref{fig:#1}}

\definecolor{anne}{rgb}{0.635,0.998,0.722}
\definecolor{stefan}{rgb}{0.998,0.722,0.635}
\definecolor{jonas}{rgb}{0.68, 0.89, 0.9}
\definecolor{liza}{rgb}{0.59, 0.44, 0.84}

\title{Negation-Instance Based Evaluation of End-to-End Negation Resolution}

\author{Elizaveta Sineva$^{1,2}$, Stefan Grünewald$^{1,2}$, Annemarie Friedrich$^1$, Jonas Kuhn$^2$\\
    $^1$Bosch Center for Artificial Intelligence, Renningen, Germany\\
    $^2$Institut für Maschinelle Sprachverarbeitung, University of Stuttgart\\
    \texttt{elisavetas2106@gmail.com}\\
    \texttt{stefan.gruenewald|annemarie.friedrich@de.bosch.com}\\
    \texttt{jonas.kuhn@ims.uni-stuttgart.de}
}

\date{}

\begin{document}
\maketitle
\begin{abstract}
In this paper, we revisit the task of negation resolution, which includes the subtasks of cue detection (e.g. \enquote{not}, \enquote{never}) and scope resolution.
In the context of previous shared tasks, a variety of evaluation metrics have been proposed.
Subsequent works usually use different subsets of these, including variations and custom implementations, rendering meaningful comparisons between systems difficult.
Examining the problem both from a linguistic perspective and from a downstream viewpoint, we here argue for a negation-instance based approach to evaluating negation resolution.
Our proposed metrics correspond to expectations over per-instance scores and hence are intuitively interpretable.
To render research comparable and to foster future work, we provide results for a set of current state-of-the-art systems for negation resolution on three English corpora, and make our implementation of the evaluation scripts publicly available.
\end{abstract}

\section{Introduction}
\label{sec:intro}

Negation is a complex semantic phenomenon in natural language that \enquote{transforms an expression into another expression whose meaning is in some way opposed to the original} \citep{morante_blanco_2021}.
It occurs frequently, with the proportion of sentences with negation in English corpora ranging between 9 and 32\% \citep{jimenez-zafra-etal-2020-corpora}.
Natural Language Processing (NLP) applications that may benefit from negation resolution include sentiment analysis \citep{wiegand-etal-2010-survey,moore-barnes-2021-multi} and information extraction.
Negation is also still a challenge in machine translation \citep{fancellu-webber-2015-translating,bentivogli-etal-2016-neural,hossain-etal-2020-non} and natural language inference \citep{hossain-etal-2020-analysis,geiger-etal-2020-neural}.

Negation Resolution \cite{morante_blanco_2021} refers to the task of automatically retrieving the elements of a sentence that are affected by the negation introduced by a \textit{cue}.
The cue's \emph{scope} is \enquote{the part of the meaning that is negated} \citep{huddleston_pullum_2002}.
The task is difficult due to the multitude of ways in which negation may be expressed.
Despite having been a continuously active research area especially since two shared tasks \citep{farkas-etal-2010-conll,morante-blanco-2012-sem}, building robust computational models is far from being a solved task, in part due to a lack of annotation standards \citep{jimenez-zafra-etal-2020-corpora}.

\begin{figure}[t]
	\footnotesize
		\fbox{%
		\hspace{-3mm}
		\renewcommand{\arraystretch}{1.2}
		\setlength{\tabcolsep}{2pt}
		\begin{tabular}{lp{41mm}cc}
		& \textbf{Test Sentences} (\hl{gold} \hl{scope}) & \dotuline{System A} & \rAB{white}{blue}{System B}\\
		\\[-0.8em]
		& & {\scriptsize TP\,/\,FP\,/\,FN} & {\scriptsize TP\,/\,FP\,/\,FN} \\
		\midrule
		(1) & \dotuline{If} \textbf{not}, \dotuline{I} \dotuline{'ll} \dotuline{have} to do with you. & 0 / 4 / 0 & 0 / 0 / 0 \\
		& & {\large\red{\xmark}} & \textcolor{darkgreen}{\large \cmark} \\
		\midrule
	    (2) & \rAB{white}{blue}{\hl{He}}  \rAB{white}{blue}{\hl{made}} \textbf{no}  \rAB{white}{blue}{\dotuline{\hl{remark}}},  but the & 1 / 0 / 2 & 3 / 0 / 0\\
	    & matter remained in his thoughts. & \red{\large \xmark} & {\large\textcolor{darkgreen}{\cmark}}\\
	    \midrule
        (3) & Well, Mrs. Warren, I can see that & 16 / 0 / 0 & 10 / 2 / 6\\
        & you have any particular cause & {\large\textcolor{darkgreen}{\cmark}} & {\large(\textcolor{darkgreen}{\cmark})}\\
        & \rAB{white}{blue}{for} \rAB{white}{blue}{concern}, \textbf{nor}  \rAB{white}{blue}{\dotuline{\hl{do}}} \rAB{white}{blue}{\dotuline{\hl{I}}} \\ &  \rAB{white}{blue}{\dotuline{\hl{understand}}}  \rAB{white}{blue}{\dotuline{\hl{why}}} \rAB{white}{blue}{\dotuline{\hl{I}}}, \dotuline{\hl{whose} \hl{time}} & \\
        &  \dotuline{\hl{is} \hl{of} \hl{some} \hl{value}}, \dotuline{\rAB{white}{blue}{\hl{should}}} & \\
        &  \dotuline{\rAB{white}{blue}{\hl{interfere}} \rAB{white}{blue}{\hl{in}} 
	    \rAB{white}{blue}{\hl{the}}  \rAB{white}{blue}{\hl{matter}}}.\\
	    \midrule
	    & \multicolumn{1}{r}{\textbf{Scope Tokens P/R:}} & 81.0/89.5 & 86.7/68.4\\
	    & \multicolumn{1}{r}{\textbf{F1:}} & \textbf{85.0} & 76.5\\
	    \midrule
	    & \multicolumn{1}{r}{\textbf{Instance-Based P/R:}} & 66.7/77.8 & 94.4/87.5\\
	    & \multicolumn{1}{r}{\textbf{F1:}} & 71.8 & \textbf{90.8}\\
	    
		\end{tabular}
		}
	\caption{\textbf{Different scoring metrics for negation scope resolution predictions.} \textcolor{darkgreen}{\cmark}/\textcolor{red}{\xmark} = our judgment of system correctness taking into account linguistic criteria. Parentheses indicate partial correctness.}
	\label{fig:teaser-img}
\end{figure}

Negation resolution has traditionally been addressed by heavily relying on syntactic parses \citep[e.g.][]{graillet2007, SohnDepParser, DEEPEN}.
Recently, end-to-end neural approaches to modeling negation resolution  \citep{fancellu-etal-2016-neural,fancellu2018neural,khandelwal-sawant-2020-negbert,kurtz-etal-2020-end} have claimed superior performance.

A significant problem in the field of negation resolution is that due to the large variety of possible evaluation setups and metrics, and the use of standard scoring scripts vs. custom implementations, it is not obvious how to meaningfully compare existing approaches and benchmark new models.
Differences in evaluation setup include, for example, whether gold cues are provided for scope resolution, and which subset and variations of evaluation metrics are used.
In the CoNLL 2010 Shared Task on detecting hedges and resolving their scope \citep{farkas-etal-2010-conll}, only exact cue and scope matches were counted.
The *SEM 2012 Shared Task on negation resolution  \citep{morante-blanco-2012-sem} proposed additional less strict metrics based on token-level matching.
The organizers of both tasks explicitly state that their evaluation is intended as a starting point and that further work on defining a scope evaluation measure that better captures the impact of partial matches is necessary.

In this paper, we take a step back and revisit the evaluation of negation resolution in order to provide a more unified picture of the relative performance of different systems.
As a result, we propose a negation-instance based evaluation framework that defines intuitively interpretable and linguistically motivated metrics, facilitating a graded scoring of cue and scope matching.
As illustrated by \fref{teaser-img}, a gradual judgment of a system's scope resolution capability may be misleading if simply computing F1 over tokens in gold or predicted scopes.
In (1), system A marked a non-existent scope, which may be detrimental to a user's trust in the system.
In (2), A did not mark the main event, i.e., extracting a correct logical representation would be impossible.
In (3), despite getting a bad recall score, B's output captures all relevant arguments and complements of the negated proposition headed by \enquote{understand.}

The core of our proposed method is to normalize precision and recall scores per instance and compute an expectation for these instance-wise match scores treating all instances equally during score aggregation, following the insight that failing on short or long negation scopes may be equally detrimental.
We apply our metrics to a set of recent neural models for negation resolution, including NegBERT, tagging-based, and dependency-parsing based approaches, providing a competitive range of baselines for future work to compare with.

Our \textbf{contributions} are as follows.
(1) We provide a concise and formal overview of existing evaluation metrics with the aim of facilitating a principled comparison of approaches to negation resolution.
(2) We propose a linguistically motivated and intuitively interpretable negation-instance based scope resolution scoring framework.
(3) Using our (and previously proposed) metrics, we conduct a reproducibility study on negation resolution, reporting performance scores of a variety of relevant baseline systems in a uniform experimental setup.\footnote{Our code is available at \url{github.com/boschresearch/negation_resolution_evaluation_conll2021}.}
(4) As a side result, we show that a modern transformer-based reimplementation of the tagging-based system by \citet{fancellu-etal-2016-neural} achieves the best negation resolution performance under most circumstances.

\section{Related Work}
\label{sec:relwork}

We here introduce the linguistic terminology and concepts used in this paper, and briefly survey related work in computational linguistics.

\subsection{Linguistic Background}
A negation \textbf{cue} signals to the listener or reader that the inverse of something is referred to.
In the ConanDoyle-neg (CD-neg) annotation scheme \citep{morante2011annotation}, a negation cue may be a single word such as \enquote{not,} a multi-word expression such as \enquote{no more,} or a negation affix such as \enquote{im} (e.g., in \enquote{imprecise}).
The \textbf{scope} is the part of the sentence that is \enquote{affected} by the negation signaled by the cue \citep{huddleston_pullum_2002}.

Logically, negation is an operator taking a proposition, which typically correspond to (sub-)clauses (e.g., $like(m,p)$ for \enquote{Mary likes pizza}), converting the corresponding assertion to an assertation stating that something is not the case ($\neg like(m,p)$ for \enquote{Mary does \textbf{not} like pizza}).
However, in natural language, depending on the embedding context as well as pragmatic presuppositions, it is not always the case that negation operators convert something to its logical complement \citep{horn2010,blanco2011some}.
For example, \enquote{she is not unhappy} does not mean \enquote{she is happy.}

In CD-neg, the aim of the annotation is to make explicit which event (process, or state) is affected by the change of polarity \citep{morante2011annotation,morante-daelemans-2012-conandoyle}.
The word referring to the event is marked as \textit{Event}, but only if the event is factual.
Thus, in cases such as \enquote{He may not know the answer,} no event is annotated. 
To avoid terminological confusion, in this paper, we ignore \textit{Event} annotations and call \enquote{know} in the example above the \textit{main predicate} of the negation.

The scope is annotated as the longest relevant part of the sentence, i.e., as the main predicate referring to the negated event and all its arguments and complements.
In contrast to BioScope \citep{szarvas-etal-2008-bioscope}, CD-neg includes the subject, but not the cue in the scope.
In constituent negation, the negation marker is attached to the object as in \enquote{Mary came to the lecture with \underline{\textbf{no} books}.}
Still, the negation scopes over the entire sentence and is marked accordingly in CD-neg to achieve representational equality with the sentence \enquote{Mary did \textbf{not} come to the lecture with books.}
A constituent-negated subject receives the same treatment.

One element of the scope is singled out as the negation's \textbf{focus}, i.e., the part that is \textit{intended} to be interpreted as false \citep{huddleston_pullum_2002}.
Detecting the focus usually requires leveraging phonetic cues as in \enquote{Your kids don't \textit{hate} school} vs. \enquote{\textit{Your} kids don't hate school} \citep{blanco2011some}.
Correct identification of a negation's focus is key to natural language understanding.
However, to date, no corpora annotating both negation scopes and focus exist.
For some ideas on integrating focus identification into our proposed evaluation framework, see \sref{discussion}.

\subsection{Automatic Negation Resolution}
Computational work on negation resolution is generally based on small- to medium-scale corpora, which in addition are often not very compatible due to differences in the employed annotation schemes and underlying tokenization.
\citet{jimenez-zafra-etal-2020-corpora} provide a comprehensive survey of datasets annotated for negation.

As the problem of negation resolution is closely tied to syntax, there are many works leveraging syntactic information, using rules over syntactic structures to resolve negation or speculation scopes \citep[e.g., ][]{velldal-etal-2012-speculation,packard-etal-2014-simple,mckenna-steedman-2020-learning} or training systems with explicit or learned syntactic features \citep{read-etal-2012-uio1,lapponi-etal-2012-uio,enger-etal-2017-open,ren2018detecting,jimenez-zafra-etal-2020-detecting}.

\citet{li-etal-2010-learning} frame scope resolution as a shallow semantic parsing task backed up by syntactic parses.
In the neural age, \citet{kurtz-etal-2020-end} frame negation resolution as a dependency parsing task.
Several works using neural networks \citep[e.g.,][]{fancellu-etal-2016-neural,fancellu2018neural,lazib2020syntactic} train BiLSTMs, or syntactically structured BiLSTMs or GCNs. %
\citet{qian-etal-2016-speculation} propose a CNN-based architecture combined with some path/position information.
Recently, a range of papers has explored BERT-based models for negation resolution \citep{khandelwal-sawant-2020-negbert,khandelwal-britto-2020-multitask,britto2020resolving,shaitarova-rinaldi-2021-negation}.
Further related work includes datasets annotated for focus \citep{,blanco-moldovan-2011-semantic,altuna-etal-2017-scope}, and the computational modeling thereof \citep[e.g.,][]{hossain-etal-2020-predicting}. 
In addition, there is a growing body of work addressing negation within the context of neural language models and commonsense reasoning using them \citep[e.g.,][]{hossain-etal-2020-analysis,geiger-etal-2020-neural,hosseini-etal-2021-understanding,jiang-etal-2021-im}.

\section{Evaluation Metrics and Settings}
\label{sec:metrics-setups}

In this section, we first give an overview of the various evaluation metrics used in shared tasks and research publications.
We then propose a framework for negation resolution focusing on instance-level metrics.
Negation resolution as a standalone NLP task is usually split into two sub-tasks, cue detection and scope resolution, where the latter depends on the former.\footnote{\citet{morante-blanco-2012-sem} also use \textit{scope resolution} to refer to the entire task of negation resolution.}
Hence, scope resolution may be evaluated in two \textbf{settings}, with either \textit{gold cue} information being given or in an end-to-end manner where systems also have to \textit{predict cues}.

We first introduce some notations with the aim of a unified presentation of metrics.
A \textbf{negation instance} is a tuple $(c, s)$ consisting of a set of cue tokens $c = (c^1, c^2, ..., c^C)$ and a set of scope tokens $s = (s^1, s^2, ..., s^S)$.
Punctuation may be excluded from these sets by definition.
In the case of affix negation, the affix is treated as a separate token if annotated or predicted as a cue. %
$I_g$ is the set of gold standard, $I_p$ the set of predicted negation instances.

\subsection{Metrics used in Shared Tasks}
Previously proposed metrics for negation resolution can be divided into metrics requiring exact cue matches and metrics requiring partial cue matches (perhaps confusingly called \textit{No Cue Match}).

\paragraph{Cue Detection.}
In this step, gold standard and predicted cue annotations are matched to each other.
When matching the gold standard cue $c_g$ and the predicted cue $c_p$, we can either require an exact match ($c_g = c_p$) or a partial match ($c_g \cap c_p \neq \varnothing$).

The CoNLL 2010 Shared Task on detecting hedges and their scope in text \citep{farkas-etal-2010-conll} required exact cue matches, while the *SEM 2012 Shared Task \citep{morante-blanco-2012-sem} employed two metrics, one counting only exact and one counting also partial cue matches as true positives.
For these metrics, each negation instance (called \textit{scope} in these works) counts as an instance, which can either be evaluated as a true positive (TP), false positive (FP), or false negative (FN).
Precision is then computed as TP/(TP+FP), recall is TP/(TP+FN), and F1 is used to summarize the scores.

\paragraph{Scope Resolution.}
The CoNLL 2010 Task employed only a single strict metric requiring exact cue and scope matches, then computing negation-instance level precision, recall and F1.
This metric is intuitive but (as the authors concede) maybe a bit too strict.
For example, the metric gives little insight when trying to identify which to two imperfect systems is slightly better at scope resolution, or when trying to evaluate whether a system under- or overpredicts the extents of scopes.
In addition, as pointed out by \citet{morante-blanco-2012-sem}, the metric penalizes partially matched negation instances more than missed instances, as these cases count both as an FN and an FP.

As a remedy, the *SEM 2012 Shared Task evaluation employs a suite of scores.
Their \textbf{Scope-Level Cue Match (SCM)} metric requires an exact cue match, while the \textbf{Scope-Level No Cue Match} metric only requires a partial cue match (but exact matching of the scope).
In both cases, partial matches are counted only as FNs; however, this results in the problem that TP+FP (the denominator in computing precision) does not correspond to the number of system predictions.
Hence, a second version called \textbf{\enquote{B}} is employed that uses the number of system predictions as the denominator.

For giving credit to partial scope matches, a so-called \textbf{Scope Tokens  (ST)} metric is used, which computes precision, recall, and F1 for tokens belonging to the scope of a negation instance in the gold data vs. system output.
Notably, one token can be counted several times if it belongs to more than one scope.\footnote{This is not stated explicitly in \citet{morante-blanco-2012-sem}, see original evaluation script: \url{https://www.clips.uantwerpen.be/sem2012-st-neg/data.html}.}
However, each cue must belong to exactly one negation instance.
Formally, the scores are computed as follows:\footnote{The notation under the sum symbol is to be read as \enquote{for each element in $I_g$, identify at most one element in $I_p$ for which the criterion $c_g = c_p$ holds.}
Replacing the requirement $c_g = c_p$ in the formulas below with $c_g \cap c_p \neq \varnothing$ results in a version using partial cue matches. However, we argue that this (a) results in technical difficulties for how to integrate partial matches, rending scores less intuitive, and (b) from an downstream point of view, it is crucial to know the full cue. Consider, for instance, the implications of \enquote{There should be \textbf{no} problems} vs. \enquote{There should be \textbf{no more} problems.}}

$\displaystyle P_{tok} = \frac{ \sum_{(c_g,s_g) \in I_g, (c_p, s_p) \in I_p, c_g = c_p} |s_g \cap s_p|}{\sum_{(c_p, s_p) \in I_p} |s_p|}$\vspace*{2mm}

$\displaystyle R_{tok} = \frac{ \sum_{(c_g,s_g) \in I_g, (c_p, s_p) \in I_p, c_g = c_p} |s_g \cap s_p|}{\sum_{(c_g, s_g) \in I_g} |s_g|}$\vspace*{2mm}

The *SEM 2012 task employs additional metrics such as an F1 over \textbf{Negated Events} (independently of cues and scopes), as well as \textbf{Global Negation} which requires cue, scope and event to be correct. %
Finally, they also report the percentage of correct negation sentences (CNS).\footnote{\citet{fancellu-etal-2017-detecting,fancellu2018neural} also report PCS, \enquote{the proportion of negation scopes that we fully and exactly match in the test corpus,} which should be the recall of the B-version of SCM.}

\subsection{Motivation for Instance-based Scoring}
From the perspective of a linguist, it matters to correctly resolve the scope pertaining to a negation cue, as this is a prerequisite for modeling the semantics of the sentence, e.g., using predicate logic.
For a machine comprehension downstream task such as sentiment analysis, it matters that a truth-theoretic interpretation of the sentence would come out correctly, or that the parts of the sentence that the model bases its decision on are interpreted using the correct polarity.
NLP tasks such as semantic parsing and event extraction care about factors similar to the predicate logic view.
In typical relation extraction setups, events are only indicated implicitly via relations between participants; here, it is of high relevance that a system includes the negated proposition's arguments in the scope.

Both the linguistic and the downstream task views desire that all parts of the negated proposition(s) are detected (high recall), but no more than these (high precision).
Ideally, a user of a system could obtain its precision and recall
in terms of (a) correctly identifying negation cues and (b) correctly identifying the negated propositions including arguments and complements by indicating the corresponding spans in the surface text.
We argue that without any prior information on the type of negation or the average scope length in the application domain given, a score computed based on existing annotated gold standard data should reflect an expectation of how well a system would perform on a random instance.

It is without question that a metric able to identify gradual differences between systems and configurations is invaluable for research and development.
The *SEM 2012 Scope Tokens F1 metric, however, effectively weights instances by their scope lengths, with longer scopes contributing more to the overall scores.
We here argue that this is not desirable.
If a system gets a fair amount of long-scope cases as in example (3) in \fref{teaser-img} right, the system will obtain a high overall score.
A system that is good at recognizing the exact extents of short scopes (which may not be trivial, especially in long sentences), will not perform on par when using such a metric.
From an application point of view, predicting a scope where none exists such as in example (1) in \fref{teaser-img} may be quite detrimental to a user's trust in the system, yet, this would have little impact on the system's score.

\subsection{Negation-Instance Based Scoring (NIS)}
\label{sec:nis-scoring}
In this section, we propose a new flexible scoring framework for negation resolution.
The aim of the scope match metric(s) is to summarize how well a negation resolution system performs overall in terms of being precise and capturing all relevant instances (recall), giving partial credit for scope resolution.
The final scores can be intuitively interpreted as the expectation how a system would perform on a random unseen instance.
For each pair of negation instances whose cues match exactly, the scope match scoring functions $f_P$ and $f_R$ compute a precision and recall score, respectively.
These scores must each range between 0 and 1.

\noindent
$\displaystyle
P_{inst} =
\displaystyle \sum_{(c_p,s_p) \in I_p, (c_g, s_g) \in I_g, c_g = c_p} \frac{1}{|I_p|} f_P(s_g, s_p)$\vspace*{3mm}

\noindent
$\displaystyle
R_{inst} = \displaystyle \sum_{(c_g,s_g) \in I_g, (c_p, s_p) \in I_p, c_g = c_p} \frac{1}{|I_g|}  f_R(s_g, s_p)
$\vspace*{3mm}

The formulas above can be interpreted as summing the scores of all gold standard / predicted instances for which a match could be found.
In other words, predicted instances for which no cue match has been found in the gold data contribute to the sum with a precision score of 0.
Similarly, in the case of recall, gold instances for which no match has been found implicitly contribute with a score of 0.

In our standard metric, giving credit to partial matches, we compute token-level scope matching scores as follows:

$\displaystyle f_{P,tok}(s_g, s_p) =
\begin{cases}
\frac{|s_g \cap s_p|}{|s_p|} & \text{if}\ |s_p| > 0\\
1 & \text{else}\\
\end{cases}
$\vspace*{3mm}

$f_{R,tok}(s_g, s_p) =
\begin{cases}
\frac{|s_g \cap s_p|}{|s_g|} & \text{if}\ |s_g| > 0\\
1 & \text{else}\\
\end{cases}
$\vspace*{3mm}

In our formulation above, we weight the match scores returned by $f_P$ or $f_r$ by $\frac{1}{|I_p|}$ or $\frac{1}{|I_g|}$, respectively. %
In other words, our final \textbf{negation-instance scores} (\textbf{\nispar}) correspond to precision and recall scores that are the expectations of the instance-level scores when weighting each instance equally, and thus allow a somewhat intuitive interpretation.

In the strictest case requiring exact scope match, the scope matching functions can be defined as:

\noindent
$\displaystyle f_{P,ex}(s_g, s_p) = f_{R,ex}(s_g, s_p) =
\begin{cases}
1 & \text{if}\ s_p = s_g\\
0 & \text{else}\\
\end{cases}$\vspace*{3mm}

In this case, our metric (\textbf{NIS\textsubscript{ex}}) would correspond to \textbf{SCM-B}, the *SEM 2012 B-style Scope-Level F1 (and to the CoNLL 2010 metric).

When comparing the *SEM 2012 Scope Tokens metric to \nispar, for computing precision, the weighting term $\frac{1}{|I_p|}$ is replaced with $\frac{|s_p|}{Z}$ ($\frac{|s_g|}{Z}$ in the case of recall).
$Z$ is the sum of all predicted (gold standard) scope tokens (see also \aref{neg-inst-formulas}).
Hence, longer scopes have a higher impact, and the resulting scores are not interpretable as expectations for a random unseen instance.
As we have argued above, this may not reflect a system's capacity of negation scope resolution in an ideal way.

\section{Modeling}
\label{sec:models}

In this section, we explain the negation resolution models that we compare in our experiments in \sref{experiments}.
Due to the large number of negation resolution systems, often with no published code, it is infeasible to provide unified evaluation scores for all prior work.
Instead, to provide competitive baselines for future work to compare with, we chose a wide range of neural architectures inspired by previous work and re-implement them.
For comparability reasons, we base them all on the same robust transformer-based language model.

\textbf{Token representation.}
Our token embedding backbone is the transformer-based XLM-R-large language model \citep{conneau-etal-2020-unsupervised}, as well as the corresponding word-piece tokenizer.
To obtain contextualized word embeddings, 
we take a weighted sum of the internal states corresponding to the first word piece for each token.
The coefficients of this weighted sum are learned during training, employing layer dropout \citep[see ][]{kondratyuk-straka-2019-75}.
Transformer weights are fine-tuned during training.
To determine the effect of injecting implicit syntactic knowledge into the system, in addition to using the default pre-trained XLM-R model, we also run experiments on an XLM-R-\textbf{synt} model that was previously fine-tuned on the task of Universal Dependencies parsing on the EWT treebank \citep{silveira-etal-2014-gold}.

\subsection{Sequence-Tagging Based Approaches}
Tagging pipelines first identify negation cues, and then for each cue, identify its scope using a second tagger.
The \textbf{NegBERT} system \citep{khandelwal-sawant-2020-negbert,britto2020resolving}, which we run using XLM-R, modifies the input for the second step, adding artificial tokens to indicate cues. %
In addition, we implement a \textbf{BiLSTM-Tagger} following the architecture proposed by \citet{fancellu-etal-2016-neural}, but using XLM-R as the underlying language model.\footnote{\citet{fancellu-etal-2016-neural} use task-specific learned or word2vec embeddings \citep{word2vec}.}
Cues are predicted using a single linear layer with softmax on top of XLM-R. Scopes are predicted by feeding, for each negation instance, the XLM-R embeddings of the sentences concatenated with a \textit{cue}/\textit{notcue} embedding to a single-layer BiLSTM and once again using a token-wise linear+softmax layer for classification.

\subsection{Dependency-Parsing Based Approaches}
This class of models frame negation resolution as a dependency parsing (\textbf{DP}) task, as proposed by \citet{kurtz-etal-2020-end}, predicting cues and scopes in a single step by encoding negation instance annotations as dependency trees.
The systems we present here differ w.r.t. this encoding (see \fref{dep-encodings}).
In the \textbf{direct mapping} \citep{kurtz-etal-2020-end}, cue tokens are modeled as dependents of the artificial root token, and scope and event tokens are attached via a dependency link to all cues they belong to.
In addition, we propose a \textbf{nested mapping} in which in the cases of embedded scopes, there is only one link from the outer scope's cue to the inner scope's cue, and all other scope tokens are only linked to their corresponding nearest cue.
We build these models using the graph-based dependency parser STEPS \citep{gruenewald2021applying}.\footnote{\url{https://github.com/boschresearch/steps-parser}}

\begin{figure}
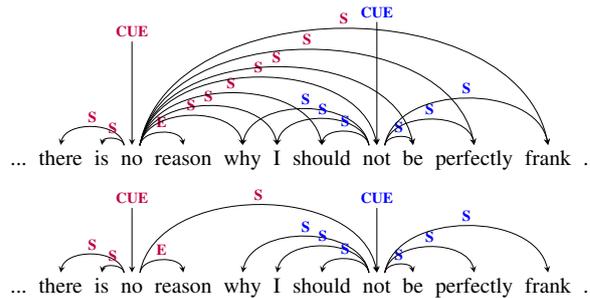

\resizebox{.5\textwidth}{!}{
\begin{dependency}[theme = simple, label style={font=\bfseries,thick}]
]
\begin{deptext}[column sep=.1em]
... \& there \& is \& no \& reason \& why \& I \& should \& not \& be \& perfectly \& frank \& . \\
\end{deptext}

\depedge{4}{2}{\color{purple}S}  %
\depedge{4}{3}{\color{purple}S} %
\deproot{4}{\color{purple}CUE}  %
\depedge{4}{5}{\color{purple}E} %
\depedge{4}{6}{\color{purple}S} %
\depedge{4}{7}{\color{purple}S} %
\depedge{4}{8}{\color{purple}S} %
\depedge{4}{9}{\color{purple}S} %
\depedge{4}{10}{\color{purple}S} %
\depedge{4}{11}{\color{purple}S} %
\depedge[edge unit distance=1.8ex]{4}{12}{\color{purple}S} %

\depedge{9}{6}{\color{blue}S} %
\depedge{9}{7}{\color{blue}S} %
\depedge{9}{8}{\color{blue}S} %
\deproot[edge unit distance=3.5ex]{9}{\color{blue}CUE} %
\depedge{9}{10}{\color{blue}S} %
\depedge{9}{11}{\color{blue}S} %
\depedge{9}{12}{\color{blue}S} %

\end{dependency}}

\resizebox{.5\textwidth}{!}{
\begin{dependency}[theme = simple, label style={font=\bfseries,thick}]
\begin{deptext}[column sep=.1em]
... \& there \& is \& no \& reason \& why \& I \& should \& not \& be \& perfectly \& frank \& . \\
\end{deptext}

\depedge{4}{2}{\color{purple}S}  %
\depedge{4}{3}{\color{purple}S} %
\deproot[edge unit distance=2ex]{4}{\color{purple}CUE}  %
\depedge{4}{5}{\color{purple}E} %
\depedge{4}{9}{\color{purple}S} %

\depedge{9}{6}{\color{blue}S} %
\depedge{9}{7}{\color{blue}S} %
\depedge{9}{8}{\color{blue}S} %
\deproot[edge unit distance=2ex]{9}{\color{blue}CUE} %
\depedge{9}{10}{\color{blue}S} %
\depedge{9}{11}{\color{blue}S} %
\depedge{9}{12}{\color{blue}S} %
	
\end{dependency}}

\caption{\textbf{Direct} (upper) vs. \textbf{nested} (lower) encoding.}
\label{fig:dep-encodings}
\end{figure}

\section{Experiments}
\label{sec:experiments}

We here report our results for end-to-end negation resolution including cue detection and scope resolution.
In all of our evaluations, we ignore punctuation tokens.
In addition to the models explained in \sref{models}, we report results for a punctuation baseline \textbf{(Punct-BL)} that uses gold cues and tags everything between the cue and the next punctuation marker as the scope.

\begin{table*}[ht]
\footnotesize
\centering
\begin{tabular}{l|r|rr|rrr|rrr}
\toprule
\textbf{}&\textbf{Cues-B}&\textbf{SCM}&\textbf{SCM-B (NIS\textsubscript{ex})}&\textbf{ST}&\textbf{ST}&\textbf{ST}&\textbf{\nispar}&\textbf{\nispar}&\textbf{\nispar}\\
\textbf{System} & F\textsubscript{1} & F\textsubscript{1} &            F\textsubscript{1} &             P &             R & F\textsubscript{1} &             P &             R & F\textsubscript{1} \\
\midrule
\textit{Punct-BL}&\textit{100.0}&\textit{17.0}&\textit{9.9}&\textit{89.8}&\textit{62.0}&\textit{73.3}&\textit{93.3}&\textit{58.8}&\textit{72.1}\\
           NegBERT &  91.1 &               78.8 &                 \textbf{70.1} &          86.1 & \textbf{87.8} &               \underline{86.9} &          83.9 & \textbf{88.7} &               \underline{86.3} \\
         BiLSTM-Tagger &               92.3 &      \textbf{79.6} &                          70.0 & \textbf{90.8} &          85.4 &      \textbf{88.0} & \textbf{90.3} &          86.3 &      \textbf{88.2} \\
      DP-direct &               92.8 &               73.2 &                          61.4 &          85.4 &          87.4 &               86.4 &          84.6 &          87.7 &               86.1 \\
      DP-nested &               93.3 &               72.8 &                          60.6 &          86.6 &          83.0 &               84.8 &          85.6 &          85.4 &               85.5 \\
 DP-direct-synt &               92.8 &               79.1 &                          69.2 &          87.4 &          86.5 &               \underline{86.9} &          86.2 &          88.3 &               \underline{87.2} \\
 DP-nested-synt &      \textbf{93.4} &               79.2 &                          69.2 &          88.0 &          84.6 &               86.2 &          87.2 &          86.6 &               86.9 \\
\bottomrule
\end{tabular}
\caption{\textbf{Comparison of Evaluation Scores} on when training and testing on \textbf{CD-neg}. For cues, we report the *SEM 2012 B version for exact cue matching \textbf{(Cues-B)}. \textbf{SCM} and \textbf{SCM-B} refer to the standard and B versions of Scope Level F1 of *SEM 2012, respectively; \textbf{ST} refers to the Scope Tokens metric. \underline{\smash{Underlined: Comparison NegBERT vs. DP-direct-synt.}}}
\label{tab:cd-results-main}
\end{table*}

\paragraph{Datasets.}
\label{sec:data}
We conduct our experiments on three corpora from a variety of domains.
In our experiment, we focus on the CD-neg (\textbf{CD}) dataset, which has been shown to be most challenging among the English negation corpora \citep{fancellu-etal-2017-detecting}. %
The corpus comprises seven literary texts corresponding to 5,520 sentences with 1,432 negation instances.
CD annotates \enquote{neither...nor} as a multiword cue with a single scope.
However, from a semantic point of view, we interpret \enquote{Neither Mary nor Sam like pizza} as $\neg like(m,p) \wedge \neg like(s,p)$, which actually suggests annotating two separate instances with overlapping scopes.\footnote{The original annotation may be seen as corresponding to an (equivalent) formalization of $\neg \left( like(m,p) \vee like(s,p) \right)$.}
We noticed that the majority of cases in which systems got multiword cue detection wrong were like this, detecting only part of the cue but resolving the scope correctly.
Rather than punishing systems for this, we decide to re-annotate the dataset accordingly, fixing a total of 10 cases.

In addition, we conduct experiments on the \textbf{BioScope} corpus \citep{szarvas-etal-2008-bioscope} using the abstracts subset (11,871 sentences), %
as well as the \textbf{SFU Review} corpus \citep{taboada-etal-2006-methods, konstantinova-etal-2012-review}, which comprises 400 reviews in eight different domains. %
The BioScope and SFU datasets are also annotated for speculation; our experiments only use the negation annotations.
More details on all corpora are given in \tref{data_negspec} in \aref{additional_figures_tables}.

\subsection{Experimental Setup}
We use the official training-dev-test split for CD-neg.
For BioScope and SFU, we create our own 80-10-10 splits for these datasets.
For more information, see \aref{additional_figures_tables}.
To tokenize BioScope, we use NLTK \cite{loper-bird-2002-nltk} with custom rules for punctuation and URLs.

Our models are implemented using PyTorch \cite{paszke2019pytorch} and the Huggingface Transformers library \cite{wolf-etal-2020-transformers}.
Training is performed on a single nVidia Tesla V100 GPU.
We use a unified set of hyperparameters for the underlying XLM-R language model, but different hyperparameters for the parser/tagger layers on top.
For a detailed description, see \aref{hyperparameters}.

\subsection{Results}
\tref{cd-results-main} shows the results for our set of competitive neural systems on CD-neg.
For cue detection, all systems perform similarly well, with the parsing-based approach using the nested representation and a syntactically fine-tuned XLM-R (DP-nested-synt) having a slight advantage.
We ran the NegBERT system for end-to-end negation resolution using the original code, but our evaluation scripts.\footnote{\citet{khandelwal-sawant-2020-negbert} report results for scope resolution that appear to be based on gold cues.}
In this unified evaluation setup, in the SCM metrics, we can only see that the DP models based on standard XLM-R underperform; all other systems perform roughly similarly.
The less strict ST and \nispar metrics reveal different precision-recall trade-offs for the direct vs. the nested encoding.
The BiLSTM-Tagger turns out to be the most accurate model for scope resolution, reflected similarly in the ST and the \nispar scores.
Our architecture similar to the one of \citet{fancellu2018neural} seems to outperform NegBERT, which adds artificial tokens to indicate cues.\footnote{We ran the scope tagger (trained on gold cues) on the output of the best cue tagger run as chosen by dev set performance. Note that while this reflects a real-life development setup, the DP models predict cues and scopes in one run. Likewise, NegBERT scores were produced by running the system off-the-shelf without optimizing the cue tagger separately.}

As expected to some extent, the \nispar and ST scores are similar in general.
Both scores identify the BiLSTM-Tagger as the most precise system and NegBERT as having the highest recall, with the Tagger achieving the best \fs.
However, while the bold-facing in \tref{cd-results-main} indicates similar patterns for the top systems, the ranking of the other systems differs when comparing ST and \nispar scores.
For example, ST assigns the same summary statistic (\fs) to NegBERT and DP-direct-synt, while in terms of \nispar, the \fs of DP-direct-synt is more than 1 point higher. %
Comparing ST and \nispar scores, we can see that \nispar generally assigns higher recall, but slightly lower precision to systems such as NegBERT or DP; the BiLSTM-Tagger's precision drops less.
Hence, the ST scores for models such as NegBERT or DP slightly under-estimate recall because the systems failed more often on longer instances; and in turn they slightly over-estimate precision, e.g., because wrongly predicted instances often have short scopes.
We here argue that while monitoring the system on several metrics is generally a good idea, \nispar constitutes a more realistic gradual end-to-end evaluation metric for negation resolution systems, and should be adopted as the main summary statistic by subsequent works or shared tasks.

\begin{table}[t]
\footnotesize
\centering
\setlength{\tabcolsep}{3pt}
\begin{tabular}{l|l|r|r|r|rrr}
\toprule
\rotatebox{90}{\textbf{}}&\rotatebox{90}{\textbf{}}&\rotatebox{90}{\textbf{Cues-B}}&\rotatebox{90}{\textbf{NIS\textsubscript{ex}}}&\rotatebox{90}{\textbf{ST}}&\rotatebox{90}{\textbf{\nispar}}&\rotatebox{90}{\textbf{\nispar}}&\rotatebox{90}{\textbf{\nispar}}\\

& \textbf{System} & F\textsubscript{1} &    F\textsubscript{1} & F\textsubscript{1} & P & R & F\textsubscript{1} \\  %
 
\midrule
\multirow{4}{*}{\rotatebox{90}{BioScope}}  & \textit{Punct-BL}&\textit{100.0}&\textit{43.2}&\textit{65.7}&\textit{77.0}&\textit{69.7}&\textit{73.5}\\
                                          &           NegBERT &          90.5 &                  79.5 &          84.5 &          83.3 &          91.5 &          87.2 \\
                                          &         BiLSTM-Tagger &          94.3 &         \textbf{83.0} &          88.6 &          91.7 &          91.7 &          91.7 \\
                                          & DP-nested-synt & \textbf{95.4} &                  82.7 & \textbf{90.4} & \textbf{92.7} & \textbf{92.1} & \textbf{92.4} \\
\midrule
\multirow{4}{*}{\rotatebox{90}{SFU}}  & \textit{Punct-BL}&\textit{100.0}&\textit{45.9}&\textit{66.5}&\textit{75.5}&\textit{87.4}&\textit{81.0}\\
                                          &           NegBERT &          81.9 &                  69.5 &          71.0 &          67.9 &          86.3 &          75.9 \\
                                          &         BiLSTM-Tagger & \textbf{86.7} &         \textbf{73.2} & \textbf{77.9} & \textbf{74.9} & \textbf{89.8} & \textbf{81.7} \\
                                          & DP-nested-synt &          86.5 &                  71.1 &          77.1 &          73.7 &          88.9 &          80.5 \\
\bottomrule
\end{tabular}
\caption{\textbf{In-Domain Comparison of Systems} on the BioScope and SFU datasets.} %
\label{tab:indomain-results-main}
\end{table}

\tref{indomain-results-main} compares the various system architectures on the BioScope and SFU data when trained and tested in-domain.
As the difference between the DP models was small (see \aref{additional_figures_tables}), we report only DP-nested-synt.
On both datasets, the DP-nested-synt models outperform NegBERT.
On BioScope, the parsing-based approach clearly outperforms NegBERT and the BiLSTM-Tagger; on SFU, the BiLSTM-Tagger performs best.

\begin{table}[t]
\footnotesize
\centering
\setlength{\tabcolsep}{4pt}
\begin{tabular}{l|ll|r|r|r|r}
\toprule
\textbf{}&\textbf{\multirow{2}{*}{\rotatebox{90}{\textbf{Train}}}}&\textbf{\multirow{2}{*}{\rotatebox{90}{\textbf{Test}}}}&\textbf{Cues-B}&\textbf{NIS\textsubscript{ex}}&\textbf{ST}&\textbf{\nispar}\\
\textbf{System} &                                                 &                                                & F\textsubscript{1} &    F\textsubscript{1} & F\textsubscript{1} & F\textsubscript{1} \\
\midrule
           NegBERT &            \multirow{3}{*}{\rotatebox{90}{Bio}} &            \multirow{3}{*}{\rotatebox{90}{CD}} &      \textbf{65.2} &                   8.9 &      \textbf{55.2} &      \textbf{51.1} \\
         BiLSTM-Tagger &                                                 &                                                &               61.6 &         \textbf{10.1} &               46.0 &               46.8 \\
 DP-nested-synt &                                                 &                                                &               64.8 &                   9.2 &               51.8 &               47.9 \\
\midrule
           NegBERT &            \multirow{3}{*}{\rotatebox{90}{SFU}} &            \multirow{3}{*}{\rotatebox{90}{CD}} &      \textbf{69.3} &                   9.7 &               56.7 &               53.6 \\
         BiLSTM-Tagger &                                                 &                                                &               68.8 &         \textbf{10.1} &      \textbf{58.0} &      \textbf{53.8} \\
 DP-nested-synt &                                                 &                                                &               68.1 &                   9.6 &               55.2 &               51.6 \\
\midrule
           NegBERT &             \multirow{3}{*}{\rotatebox{90}{CD}} &           \multirow{3}{*}{\rotatebox{90}{Bio}} &               60.6 &         \textbf{19.0} &               49.3 &               46.2 \\
         BiLSTM-Tagger &                                                 &                                                &               64.3 &                  17.7 &      \textbf{56.7} &      \textbf{54.3} \\
 DP-nested-synt &                                                 &                                                &      \textbf{65.3} &                  18.7 &               54.7 &               50.9 \\
\midrule
           NegBERT &            \multirow{3}{*}{\rotatebox{90}{SFU}} &           \multirow{3}{*}{\rotatebox{90}{Bio}} &               78.7 &                  55.5 &               63.2 &               71.0 \\
         BiLSTM-Tagger &                                                 &                                                &               81.0 &                  53.4 &               65.9 &               72.8 \\
 DP-nested-synt &                                                 &                                                &      \textbf{82.0} &         \textbf{57.2} &      \textbf{67.2} &      \textbf{74.0} \\
\midrule
           NegBERT &             \multirow{3}{*}{\rotatebox{90}{CD}} &           \multirow{3}{*}{\rotatebox{90}{SFU}} &               51.3 &                  10.0 &               37.2 &               37.5 \\
         BiLSTM-Tagger &                                                 &                                                &      \textbf{50.2} &          \textbf{9.0} &      \textbf{37.5} &      \textbf{37.8} \\
 DP-nested-synt &                                                 &                                                &               50.3 &                   9.0 &               36.2 &               36.5 \\
\midrule
           NegBERT &            \multirow{3}{*}{\rotatebox{90}{Bio}} &           \multirow{3}{*}{\rotatebox{90}{SFU}} &               63.4 &                  48.8 &               53.7 &               58.6 \\
         BiLSTM-Tagger &                                                 &                                                &      \textbf{64.5} &         \textbf{53.1} &      \textbf{57.0} &      \textbf{61.4} \\
 DP-nested-synt &                                                 &                                                &               56.9 &                  40.1 &               47.5 &               51.5 \\
\bottomrule
\end{tabular}
\caption{\textbf{Cross-Domain Comparison of Systems} between CD-neg, BioScope, and SFU.} %
\label{tab:crossdomain-results-main}
\end{table}

Finally, we also give results for cross-domain performance.
Overall, the BiLSTM-Tagger seems most robust.
However, the NegBERT system performs close to or better than the BiLSTM-Tagger when moving from another to the ConanDoyle-neg dataset, and the DP-nested-synt model has an advantage when moving from SFU to BioScope.
In sum, particularly cross-domain negation resolution is still far from being solved. We hope that our linguistically motivated evaluation framework can aid the development of more robust negation resolution systems.

\section{Discussion}
\label{sec:discussion}
A general problem with most existing text corpora annotated for negation is that annotations are only created on the surface level.
However, in the words of \citet{blanco-moldovan-2011-semantic}, \enquote{Negation does not stand on its own, to be useful, it should be added as part of another existing knowledge representation.}
Our negation-instance based framework was detailed above such that it can directly be applied to existing widely used negation corpora.
However, the precision and recall scoring functions $f_P$ and $f_R$ can easily be designed in other ways.
For instance, if leveraging a dependency parse of the sentence, argument structure could be approximated by taking as the set of elements in a scope not the tokens, but only the dependents of the main predicate to which the negation cue attaches.
In this way, relative clauses as in (3) in \fref{teaser-img} would have zero impact.
Thinking ahead, a truly linguistically motivated scoring function could even weight components by their importance of being detected as belonging to a scope, e.g., missing a restrictive relative clause could be penalized more than missing a non-restrictive one.

If we had a dataset marking cues, scopes and foci, our scoring framework could assign a high weight for detecting the focus correctly.
Similarly, detecting the main predicates correctly could be incorporated into the score.
We here decided against including event annotations as they are only marked on factual events and hence, in our opinion, should be evaluated as a separate task (as was done in most metrics in previous shared tasks).

\section{Conclusion and Outlook}
\label{sec:conclusion}
The aim of this paper is to provide a concise reference of evaluation metrics and setups for negation resolution, making it easier for NLP researchers and developers to enter the research area. %
Our core contribution is to detail the linguistic motivation for employing a new instance-based approach to evaluating the performance of end-to-end negation resolution systems, giving credit to partial scope matches but relying on exact cue matches.
We argue that this metric is well-motivated and intuitively interpretable and should hence be adopted by future studies or shared tasks.
In addition, our experimental study, comparing a set of recent neural architectures on a similar basis, will serve as a reference for future work.

Besides implementing a variety of linguistically motivated extensions with the aim of deeper system analyses using our framework as suggested above, an important next step is to evaluate the suite of models used in this paper on further datasets in languages other than English \citep[e.g.,][]{zou-etal-2015-negation,liu-etal-2018-negpar,jimenez-zafra-etal-2018-review}.

\bibliographystyle{acl_natbib}
\bibliography{references,anthology}

\begin{thebibliography}{58}
\expandafter\ifx\csname natexlab\endcsname\relax\def\natexlab#1{#1}\fi

\bibitem[{Altuna et~al.(2017)Altuna, Minard, and
  Speranza}]{altuna-etal-2017-scope}
Bego{\~n}a Altuna, Anne-Lyse Minard, and Manuela Speranza. 2017.
\newblock \href {https://doi.org/10.18653/v1/W17-1806} {The scope and focus of
  negation: A complete annotation framework for {I}talian}.
\newblock In \emph{Proceedings of the Workshop Computational Semantics Beyond
  Events and Roles}, pages 34--42, Valencia, Spain. Association for
  Computational Linguistics.

\bibitem[{Bentivogli et~al.(2016)Bentivogli, Bisazza, Cettolo, and
  Federico}]{bentivogli-etal-2016-neural}
Luisa Bentivogli, Arianna Bisazza, Mauro Cettolo, and Marcello Federico. 2016.
\newblock \href {https://doi.org/10.18653/v1/D16-1025} {Neural versus
  phrase-based machine translation quality: a case study}.
\newblock In \emph{Proceedings of the 2016 Conference on Empirical Methods in
  Natural Language Processing}, pages 257--267, Austin, Texas. Association for
  Computational Linguistics.

\bibitem[{Blanco and
  Moldovan(2011{\natexlab{a}})}]{blanco-moldovan-2011-semantic}
Eduardo Blanco and Dan Moldovan. 2011{\natexlab{a}}.
\newblock \href {https://www.aclweb.org/anthology/P11-1059} {Semantic
  representation of negation using focus detection}.
\newblock In \emph{Proceedings of the 49th Annual Meeting of the Association
  for Computational Linguistics: Human Language Technologies}, pages 581--589,
  Portland, Oregon, USA. Association for Computational Linguistics.

\bibitem[{Blanco and Moldovan(2011{\natexlab{b}})}]{blanco2011some}
Eduardo Blanco and Dan Moldovan. 2011{\natexlab{b}}.
\newblock \href
  {https://doi.org/https://aaai.org/ocs/index.php/FLAIRS/FLAIRS11/paper/view/2629/3031}
  {Some issues on detecting negation from text}.
\newblock In \emph{Twenty-Fourth International FLAIRS Conference}.

\bibitem[{Britto and Khandelwal(2020)}]{britto2020resolving}
Benita~Kathleen Britto and Aditya Khandelwal. 2020.
\newblock \href {https://doi.org/https://arxiv.org/abs/2001.02885} {Resolving
  the scope of speculation and negation using transformer-based architectures}.
\newblock \emph{arXiv preprint arXiv:2001.02885}.

\bibitem[{Conneau et~al.(2020)Conneau, Khandelwal, Goyal, Chaudhary, Wenzek,
  Guzm{\'a}n, Grave, Ott, Zettlemoyer, and
  Stoyanov}]{conneau-etal-2020-unsupervised}
Alexis Conneau, Kartikay Khandelwal, Naman Goyal, Vishrav Chaudhary, Guillaume
  Wenzek, Francisco Guzm{\'a}n, Edouard Grave, Myle Ott, Luke Zettlemoyer, and
  Veselin Stoyanov. 2020.
\newblock \href {https://doi.org/10.18653/v1/2020.acl-main.747} {Unsupervised
  cross-lingual representation learning at scale}.
\newblock In \emph{Proceedings of the 58th Annual Meeting of the Association
  for Computational Linguistics}, pages 8440--8451, Online. Association for
  Computational Linguistics.

\bibitem[{Dozat and Manning(2018)}]{dozat-manning-2018-simpler}
Timothy Dozat and Christopher~D. Manning. 2018.
\newblock \href {https://doi.org/10.18653/v1/P18-2077} {Simpler but more
  accurate semantic dependency parsing}.
\newblock In \emph{Proceedings of the 56th Annual Meeting of the Association
  for Computational Linguistics (Volume 2: Short Papers)}, pages 484--490,
  Melbourne, Australia. Association for Computational Linguistics.

\bibitem[{Enger et~al.(2017)Enger, Velldal, and
  {\O}vrelid}]{enger-etal-2017-open}
Martine Enger, Erik Velldal, and Lilja {\O}vrelid. 2017.
\newblock \href {https://doi.org/10.18653/v1/W17-1810} {An open-source tool for
  negation detection: a maximum-margin approach}.
\newblock In \emph{Proceedings of the Workshop Computational Semantics Beyond
  Events and Roles}, pages 64--69, Valencia, Spain. Association for
  Computational Linguistics.

\bibitem[{Fancellu et~al.(2016)Fancellu, Lopez, and
  Webber}]{fancellu-etal-2016-neural}
Federico Fancellu, Adam Lopez, and Bonnie Webber. 2016.
\newblock \href {https://doi.org/10.18653/v1/P16-1047} {Neural networks for
  negation scope detection}.
\newblock In \emph{Proceedings of the 54th Annual Meeting of the Association
  for Computational Linguistics (Volume 1: Long Papers)}, pages 495--504,
  Berlin, Germany. Association for Computational Linguistics.

\bibitem[{Fancellu et~al.(2017)Fancellu, Lopez, Webber, and
  He}]{fancellu-etal-2017-detecting}
Federico Fancellu, Adam Lopez, Bonnie Webber, and Hangfeng He. 2017.
\newblock \href {https://www.aclweb.org/anthology/E17-2010} {Detecting negation
  scope is easy, except when it isn{'}t}.
\newblock In \emph{Proceedings of the 15th Conference of the {E}uropean Chapter
  of the Association for Computational Linguistics: Volume 2, Short Papers},
  pages 58--63, Valencia, Spain. Association for Computational Linguistics.

\bibitem[{Fancellu et~al.(2018)Fancellu, Lopez, and
  Webber}]{fancellu2018neural}
Federico Fancellu, Adam Lopez, and Bonnie~L. Webber. 2018.
\newblock \href {http://arxiv.org/abs/1810.02156} {Neural networks for
  cross-lingual negation scope detection}.
\newblock \emph{CoRR}, abs/1810.02156.

\bibitem[{Fancellu and Webber(2015)}]{fancellu-webber-2015-translating}
Federico Fancellu and Bonnie Webber. 2015.
\newblock \href {https://doi.org/10.3115/v1/W15-1003} {Translating negation:
  Induction, search and model errors}.
\newblock In \emph{Proceedings of the Ninth Workshop on Syntax, Semantics and
  Structure in Statistical Translation}, pages 21--29, Denver, Colorado, USA.
  Association for Computational Linguistics.

\bibitem[{Farkas et~al.(2010)Farkas, Vincze, M{\'o}ra, Csirik, and
  Szarvas}]{farkas-etal-2010-conll}
Rich{\'a}rd Farkas, Veronika Vincze, Gy{\"o}rgy M{\'o}ra, J{\'a}nos Csirik, and
  Gy{\"o}rgy Szarvas. 2010.
\newblock \href {https://www.aclweb.org/anthology/W10-3001} {The {C}o{NLL}-2010
  shared task: Learning to detect hedges and their scope in natural language
  text}.
\newblock In \emph{Proceedings of the Fourteenth Conference on Computational
  Natural Language Learning {--} Shared Task}, pages 1--12, Uppsala, Sweden.
  Association for Computational Linguistics.

\bibitem[{Geiger et~al.(2020)Geiger, Richardson, and
  Potts}]{geiger-etal-2020-neural}
Atticus Geiger, Kyle Richardson, and Christopher Potts. 2020.
\newblock \href {https://doi.org/10.18653/v1/2020.blackboxnlp-1.16} {Neural
  natural language inference models partially embed theories of lexical
  entailment and negation}.
\newblock In \emph{Proceedings of the Third BlackboxNLP Workshop on Analyzing
  and Interpreting Neural Networks for NLP}, pages 163--173, Online.
  Association for Computational Linguistics.

\bibitem[{Gr{\"{u}}newald et~al.(2020)Gr{\"{u}}newald, Friedrich, and
  Kuhn}]{gruenewald2021applying}
Stefan Gr{\"{u}}newald, Annemarie Friedrich, and Jonas Kuhn. 2020.
\newblock \href {http://arxiv.org/abs/2010.12699} {Graph-based universal
  dependency parsing in the age of the transformer: What works, and what
  doesn't}.
\newblock \emph{CoRR}, abs/2010.12699.

\bibitem[{Horn(2010)}]{horn2010}
Laurence~R. Horn, editor. 2010.
\newblock \href {https://doi.org/doi:10.1515/9783110219302} {\emph{The
  Expression of Negation}}.
\newblock De Gruyter Mouton.

\bibitem[{Hossain et~al.(2020{\natexlab{a}})Hossain, Anastasopoulos, Blanco,
  and Palmer}]{hossain-etal-2020-non}
Md~Mosharaf Hossain, Antonios Anastasopoulos, Eduardo Blanco, and Alexis
  Palmer. 2020{\natexlab{a}}.
\newblock \href {https://doi.org/10.18653/v1/2020.findings-emnlp.345} {It{'}s
  not a non-issue: Negation as a source of error in machine translation}.
\newblock In \emph{Findings of the Association for Computational Linguistics:
  EMNLP 2020}, pages 3869--3885, Online. Association for Computational
  Linguistics.

\bibitem[{Hossain et~al.(2020{\natexlab{b}})Hossain, Hamilton, Palmer, and
  Blanco}]{hossain-etal-2020-predicting}
Md~Mosharaf Hossain, Kathleen Hamilton, Alexis Palmer, and Eduardo Blanco.
  2020{\natexlab{b}}.
\newblock \href {https://doi.org/10.18653/v1/2020.acl-main.743} {Predicting the
  focus of negation: Model and error analysis}.
\newblock In \emph{Proceedings of the 58th Annual Meeting of the Association
  for Computational Linguistics}, pages 8389--8401, Online. Association for
  Computational Linguistics.

\bibitem[{Hossain et~al.(2020{\natexlab{c}})Hossain, Kovatchev, Dutta, Kao,
  Wei, and Blanco}]{hossain-etal-2020-analysis}
Md~Mosharaf Hossain, Venelin Kovatchev, Pranoy Dutta, Tiffany Kao, Elizabeth
  Wei, and Eduardo Blanco. 2020{\natexlab{c}}.
\newblock \href {https://doi.org/10.18653/v1/2020.emnlp-main.732} {An analysis
  of natural language inference benchmarks through the lens of negation}.
\newblock In \emph{Proceedings of the 2020 Conference on Empirical Methods in
  Natural Language Processing (EMNLP)}, pages 9106--9118, Online. Association
  for Computational Linguistics.

\bibitem[{Hosseini et~al.(2021)Hosseini, Reddy, Bahdanau, Hjelm, Sordoni, and
  Courville}]{hosseini-etal-2021-understanding}
Arian Hosseini, Siva Reddy, Dzmitry Bahdanau, R~Devon Hjelm, Alessandro
  Sordoni, and Aaron Courville. 2021.
\newblock \href {https://www.aclweb.org/anthology/2021.naacl-main.102}
  {Understanding by understanding not: Modeling negation in language models}.
\newblock In \emph{Proceedings of the 2021 Conference of the North American
  Chapter of the Association for Computational Linguistics: Human Language
  Technologies}, pages 1301--1312, Online. Association for Computational
  Linguistics.

\bibitem[{Huddleston and Pullum(2002)}]{huddleston_pullum_2002}
Rodney Huddleston and Geoffrey~K. Pullum. 2002.
\newblock \href {https://doi.org/10.1017/9781316423530} {\emph{The Cambridge
  Grammar of the English Language}}.
\newblock Cambridge University Press.

\bibitem[{Jiang et~al.(2021)Jiang, Bosselut, Bhagavatula, and
  Choi}]{jiang-etal-2021-im}
Liwei Jiang, Antoine Bosselut, Chandra Bhagavatula, and Yejin Choi. 2021.
\newblock \href {https://www.aclweb.org/anthology/2021.naacl-main.346}
  {{``}{I}{'}m not mad{''}: Commonsense implications of negation and
  contradiction}.
\newblock In \emph{Proceedings of the 2021 Conference of the North American
  Chapter of the Association for Computational Linguistics: Human Language
  Technologies}, pages 4380--4397, Online. Association for Computational
  Linguistics.

\bibitem[{Jim{\'e}nez-Zafra et~al.(2020{\natexlab{a}})Jim{\'e}nez-Zafra,
  Morante, Blanco, Mart{\'\i}n~Valdivia, and
  Ure{\~n}a~L{\'o}pez}]{jimenez-zafra-etal-2020-detecting}
Salud~Mar{\'\i}a Jim{\'e}nez-Zafra, Roser Morante, Eduardo Blanco,
  Mar{\'\i}a~Teresa Mart{\'\i}n~Valdivia, and L.~Alfonso Ure{\~n}a~L{\'o}pez.
  2020{\natexlab{a}}.
\newblock \href {https://www.aclweb.org/anthology/2020.lrec-1.853} {Detecting
  negation cues and scopes in {S}panish}.
\newblock In \emph{Proceedings of the 12th Language Resources and Evaluation
  Conference}, pages 6902--6911, Marseille, France. European Language Resources
  Association.

\bibitem[{Jim{\'e}nez-Zafra et~al.(2018)Jim{\'e}nez-Zafra, Morante, Martin, and
  Ure{\~n}a-L{\'o}pez}]{jimenez-zafra-etal-2018-review}
Salud~Mar{\'\i}a Jim{\'e}nez-Zafra, Roser Morante, Maite Martin, and L.~Alfonso
  Ure{\~n}a-L{\'o}pez. 2018.
\newblock \href {https://www.aclweb.org/anthology/C18-1078} {A review of
  {S}panish corpora annotated with negation}.
\newblock In \emph{Proceedings of the 27th International Conference on
  Computational Linguistics}, pages 915--924, Santa Fe, New Mexico, USA.
  Association for Computational Linguistics.

\bibitem[{Jim{\'e}nez-Zafra et~al.(2020{\natexlab{b}})Jim{\'e}nez-Zafra,
  Morante, Mart{\'\i}n-Valdivia, and
  Ure{\~n}a-L{\'o}pez}]{jimenez-zafra-etal-2020-corpora}
Salud~Mar{\'\i}a Jim{\'e}nez-Zafra, Roser Morante, Mar{\'\i}a~Teresa
  Mart{\'\i}n-Valdivia, and L.~Alfonso Ure{\~n}a-L{\'o}pez. 2020{\natexlab{b}}.
\newblock \href {https://doi.org/10.1162/coli_a_00371} {Corpora annotated with
  negation: An overview}.
\newblock \emph{Computational Linguistics}, 46(1):1--52.

\bibitem[{Khandelwal and Britto(2020)}]{khandelwal-britto-2020-multitask}
Aditya Khandelwal and Benita~Kathleen Britto. 2020.
\newblock \href {https://doi.org/10.18653/v1/2020.louhi-1.9} {Multitask
  learning of negation and speculation using transformers}.
\newblock In \emph{Proceedings of the 11th International Workshop on Health
  Text Mining and Information Analysis}, pages 79--87, Online. Association for
  Computational Linguistics.

\bibitem[{Khandelwal and Sawant(2020)}]{khandelwal-sawant-2020-negbert}
Aditya Khandelwal and Suraj Sawant. 2020.
\newblock \href {https://www.aclweb.org/anthology/2020.lrec-1.704}
  {{N}eg{BERT}: A transfer learning approach for negation detection and scope
  resolution}.
\newblock In \emph{Proceedings of the 12th Language Resources and Evaluation
  Conference}, pages 5739--5748, Marseille, France. European Language Resources
  Association.

\bibitem[{Kondratyuk and Straka(2019)}]{kondratyuk-straka-2019-75}
Dan Kondratyuk and Milan Straka. 2019.
\newblock \href {https://doi.org/10.18653/v1/D19-1279} {75 languages, 1 model:
  Parsing {U}niversal {D}ependencies universally}.
\newblock In \emph{Proceedings of the 2019 Conference on Empirical Methods in
  Natural Language Processing and the 9th International Joint Conference on
  Natural Language Processing (EMNLP-IJCNLP)}, pages 2779--2795, Hong Kong,
  China. Association for Computational Linguistics.

\bibitem[{Konstantinova et~al.(2012)Konstantinova, de~Sousa, Cruz, Ma{\~n}a,
  Taboada, and Mitkov}]{konstantinova-etal-2012-review}
Natalia Konstantinova, Sheila~C.M. de~Sousa, Noa~P. Cruz, Manuel~J. Ma{\~n}a,
  Maite Taboada, and Ruslan Mitkov. 2012.
\newblock \href
  {http://www.lrec-conf.org/proceedings/lrec2012/pdf/533_Paper.pdf} {A review
  corpus annotated for negation, speculation and their scope}.
\newblock In \emph{Proceedings of the Eighth International Conference on
  Language Resources and Evaluation ({LREC}'12)}, pages 3190--3195, Istanbul,
  Turkey. European Language Resources Association (ELRA).

\bibitem[{Kurtz et~al.(2020)Kurtz, Oepen, and Kuhlmann}]{kurtz-etal-2020-end}
Robin Kurtz, Stephan Oepen, and Marco Kuhlmann. 2020.
\newblock \href {https://doi.org/10.18653/v1/2020.iwpt-1.3} {End-to-end
  negation resolution as graph parsing}.
\newblock In \emph{Proceedings of the 16th International Conference on Parsing
  Technologies and the IWPT 2020 Shared Task on Parsing into Enhanced Universal
  Dependencies}, pages 14--24, Online. Association for Computational
  Linguistics.

\bibitem[{Lapponi et~al.(2012)Lapponi, Velldal, {\O}vrelid, and
  Read}]{lapponi-etal-2012-uio}
Emanuele Lapponi, Erik Velldal, Lilja {\O}vrelid, and Jonathon Read. 2012.
\newblock \href {https://www.aclweb.org/anthology/S12-1042} {{U}i{O} 2:
  Sequence-labeling negation using dependency features}.
\newblock In \emph{*{SEM} 2012: The First Joint Conference on Lexical and
  Computational Semantics {--} Volume 1: Proceedings of the main conference and
  the shared task, and Volume 2: Proceedings of the Sixth International
  Workshop on Semantic Evaluation ({S}em{E}val 2012)}, pages 319--327,
  Montr{\'e}al, Canada. Association for Computational Linguistics.

\bibitem[{Lazib et~al.(2020)Lazib, Qin, Zhao, Zhang, and
  Liu}]{lazib2020syntactic}
Lydia Lazib, Bing Qin, Yanyan Zhao, Weinan Zhang, and Ting Liu. 2020.
\newblock \href
  {https://doi.org/https://link.springer.com/article/10.1007/s11704-018-7368-6}
  {A syntactic path-based hybrid neural network for negation scope detection}.
\newblock \emph{Frontiers of computer science}, 14(1):84--94.

\bibitem[{Li et~al.(2010)Li, Zhou, Wang, and Zhu}]{li-etal-2010-learning}
Junhui Li, Guodong Zhou, Hongling Wang, and Qiaoming Zhu. 2010.
\newblock \href {https://www.aclweb.org/anthology/C10-1076} {Learning the scope
  of negation via shallow semantic parsing}.
\newblock In \emph{Proceedings of the 23rd International Conference on
  Computational Linguistics (Coling 2010)}, pages 671--679, Beijing, China.
  Coling 2010 Organizing Committee.

\bibitem[{Liu et~al.(2018)Liu, Fancellu, and Webber}]{liu-etal-2018-negpar}
Qianchu Liu, Federico Fancellu, and Bonnie Webber. 2018.
\newblock \href {https://www.aclweb.org/anthology/L18-1547} {{N}eg{P}ar: A
  parallel corpus annotated for negation}.
\newblock In \emph{Proceedings of the Eleventh International Conference on
  Language Resources and Evaluation ({LREC} 2018)}, Miyazaki, Japan. European
  Language Resources Association (ELRA).

\bibitem[{Loper and Bird(2002)}]{loper-bird-2002-nltk}
Edward Loper and Steven Bird. 2002.
\newblock \href {https://doi.org/10.3115/1118108.1118117} {{NLTK}: The natural
  language toolkit}.
\newblock In \emph{Proceedings of the {ACL}-02 Workshop on Effective Tools and
  Methodologies for Teaching Natural Language Processing and Computational
  Linguistics}, pages 63--70, Philadelphia, Pennsylvania, USA. Association for
  Computational Linguistics.

\bibitem[{McKenna and Steedman(2020)}]{mckenna-steedman-2020-learning}
Nick McKenna and Mark Steedman. 2020.
\newblock \href {https://www.aclweb.org/anthology/2020.starsem-1.15} {Learning
  negation scope from syntactic structure}.
\newblock In \emph{Proceedings of the Ninth Joint Conference on Lexical and
  Computational Semantics}, pages 137--142, Barcelona, Spain (Online).
  Association for Computational Linguistics.

\bibitem[{Mehrabi et~al.(2015)Mehrabi, Krishnan, Sohn, Roch, Schmidt,
  Kesterson, Beesley, Dexter, Max~Schmidt, Liu, and Palakal}]{DEEPEN}
Saeed Mehrabi, Anand Krishnan, Sunghwan Sohn, Alexandra~M Roch, Heidi Schmidt,
  Joe Kesterson, Chris Beesley, Paul Dexter, C~Max~Schmidt, Hongfang Liu, and
  Mathew Palakal. 2015.
\newblock \href {https://doi.org/10.1016/j.jbi.2015.02.010} {Deepen: A negation
  detection system for clinical text incorporating dependency relation into
  negex}.
\newblock \emph{Journal of biomedical informatics}, 54:213—219.

\bibitem[{Mikolov et~al.(2013)Mikolov, Chen, Corrado, and Dean}]{word2vec}
Tom{\'{a}}s Mikolov, Kai Chen, Greg Corrado, and Jeffrey Dean. 2013.
\newblock \href {http://arxiv.org/abs/1301.3781} {Efficient estimation of word
  representations in vector space}.
\newblock In \emph{1st International Conference on Learning Representations,
  {ICLR} 2013, Scottsdale, Arizona, USA, May 2-4, 2013, Workshop Track
  Proceedings}.

\bibitem[{Moore and Barnes(2021)}]{moore-barnes-2021-multi}
Andrew Moore and Jeremy Barnes. 2021.
\newblock \href {https://www.aclweb.org/anthology/2021.naacl-main.227}
  {Multi-task learning of negation and speculation for targeted sentiment
  classification}.
\newblock In \emph{Proceedings of the 2021 Conference of the North American
  Chapter of the Association for Computational Linguistics: Human Language
  Technologies}, pages 2838--2869, Online. Association for Computational
  Linguistics.

\bibitem[{Morante and Blanco(2012)}]{morante-blanco-2012-sem}
Roser Morante and Eduardo Blanco. 2012.
\newblock \href {https://www.aclweb.org/anthology/S12-1035} {*{SEM} 2012 shared
  task: Resolving the scope and focus of negation}.
\newblock In \emph{*{SEM} 2012: The First Joint Conference on Lexical and
  Computational Semantics {--} Volume 1: Proceedings of the main conference and
  the shared task, and Volume 2: Proceedings of the Sixth International
  Workshop on Semantic Evaluation ({S}em{E}val 2012)}, pages 265--274,
  Montr{\'e}al, Canada. Association for Computational Linguistics.

\bibitem[{Morante and Blanco(2021)}]{morante_blanco_2021}
Roser Morante and Eduardo Blanco. 2021.
\newblock \href {https://doi.org/10.1017/S1351324920000534} {Recent advances in
  processing negation}.
\newblock \emph{Natural Language Engineering}, 27(2):121–130.

\bibitem[{Morante and Daelemans(2012)}]{morante-daelemans-2012-conandoyle}
Roser Morante and Walter Daelemans. 2012.
\newblock \href
  {http://www.lrec-conf.org/proceedings/lrec2012/pdf/221_Paper.pdf}
  {{C}onan{D}oyle-neg: Annotation of negation cues and their scope in conan
  doyle stories}.
\newblock In \emph{Proceedings of the Eighth International Conference on
  Language Resources and Evaluation ({LREC}'12)}, pages 1563--1568, Istanbul,
  Turkey. European Language Resources Association (ELRA).

\bibitem[{Morante et~al.(2011)Morante, Schrauwen, and
  Daelemans}]{morante2011annotation}
Roser Morante, Sarah Schrauwen, and Walter Daelemans. 2011.
\newblock Annotation of negation cues and their scope: Guidelines v1.
\newblock \emph{Computational linguistics and psycholinguistics technical
  report series, CTRS-003}, pages 1--42.

\bibitem[{Packard et~al.(2014)Packard, Bender, Read, Oepen, and
  Dridan}]{packard-etal-2014-simple}
Woodley Packard, Emily~M. Bender, Jonathon Read, Stephan Oepen, and Rebecca
  Dridan. 2014.
\newblock \href {https://doi.org/10.3115/v1/P14-1007} {Simple negation scope
  resolution through deep parsing: A semantic solution to a semantic problem}.
\newblock In \emph{Proceedings of the 52nd Annual Meeting of the Association
  for Computational Linguistics (Volume 1: Long Papers)}, pages 69--78,
  Baltimore, Maryland. Association for Computational Linguistics.

\bibitem[{Paszke et~al.(2019)Paszke, Gross, Massa, Lerer, Bradbury, Chanan,
  Killeen, Lin, Gimelshein, Antiga, Desmaison, Kopf, Yang, DeVito, Raison,
  Tejani, Chilamkurthy, Steiner, Fang, Bai, and Chintala}]{paszke2019pytorch}
Adam Paszke, Sam Gross, Francisco Massa, Adam Lerer, James Bradbury, Gregory
  Chanan, Trevor Killeen, Zeming Lin, Natalia Gimelshein, Luca Antiga, Alban
  Desmaison, Andreas Kopf, Edward Yang, Zachary DeVito, Martin Raison, Alykhan
  Tejani, Sasank Chilamkurthy, Benoit Steiner, Lu~Fang, Junjie Bai, and Soumith
  Chintala. 2019.
\newblock \href
  {http://papers.neurips.cc/paper/9015-pytorch-an-imperative-style-high-performance-deep-learning-library.pdf}
  {Pytorch: An imperative style, high-performance deep learning library}.
\newblock In H.~Wallach, H.~Larochelle, A.~Beygelzimer, F.~d'~Alch\'{e}-Buc,
  E.~Fox, and R.~Garnett, editors, \emph{Advances in Neural Information
  Processing Systems 32}, pages 8024--8035. Curran Associates, Inc.

\bibitem[{Qian et~al.(2016)Qian, Li, Zhu, Zhou, Luo, and
  Luo}]{qian-etal-2016-speculation}
Zhong Qian, Peifeng Li, Qiaoming Zhu, Guodong Zhou, Zhunchen Luo, and Wei Luo.
  2016.
\newblock \href {https://doi.org/10.18653/v1/D16-1078} {Speculation and
  negation scope detection via convolutional neural networks}.
\newblock In \emph{Proceedings of the 2016 Conference on Empirical Methods in
  Natural Language Processing}, pages 815--825, Austin, Texas. Association for
  Computational Linguistics.

\bibitem[{Read et~al.(2012)Read, Velldal, {\O}vrelid, and
  Oepen}]{read-etal-2012-uio1}
Jonathon Read, Erik Velldal, Lilja {\O}vrelid, and Stephan Oepen. 2012.
\newblock \href {https://www.aclweb.org/anthology/S12-1041} {{U}i{O}1:
  Constituent-based discriminative ranking for negation resolution}.
\newblock In \emph{*{SEM} 2012: The First Joint Conference on Lexical and
  Computational Semantics {--} Volume 1: Proceedings of the main conference and
  the shared task, and Volume 2: Proceedings of the Sixth International
  Workshop on Semantic Evaluation ({S}em{E}val 2012)}, pages 310--318,
  Montr{\'e}al, Canada. Association for Computational Linguistics.

\bibitem[{Ren et~al.(2018)Ren, Fei, and Peng}]{ren2018detecting}
Y.~Ren, H.~Fei, and Q.~Peng. 2018.
\newblock \href {https://doi.org/10.1109/BIBM.2018.8621261} {Detecting the
  scope of negation and speculation in biomedical texts by using recursive
  neural network}.
\newblock In \emph{2018 IEEE International Conference on Bioinformatics and
  Biomedicine (BIBM)}, pages 739--742, Los Alamitos, CA, USA. IEEE Computer
  Society.

\bibitem[{Sanchez~Graillet and Poesio(2007)}]{graillet2007}
Olivia Sanchez~Graillet and Massimo Poesio. 2007.
\newblock \href {https://doi.org/10.1093/bioinformatics/btm184} {Negation of
  protein protein interactions: Analysis and extraction}.
\newblock \emph{Bioinformatics (Oxford, England)}, 23:i424--32.

\bibitem[{Shaitarova and Rinaldi(2021)}]{shaitarova-rinaldi-2021-negation}
Anastassia Shaitarova and Fabio Rinaldi. 2021.
\newblock \href {https://www.aclweb.org/anthology/2021.naacl-srw.3} {Negation
  typology and general representation models for cross-lingual zero-shot
  negation scope resolution in {R}ussian, {F}rench, and {S}panish.}
\newblock In \emph{Proceedings of the 2021 Conference of the North American
  Chapter of the Association for Computational Linguistics: Student Research
  Workshop}, pages 15--23, Online. Association for Computational Linguistics.

\bibitem[{Silveira et~al.(2014)Silveira, Dozat, de~Marneffe, Bowman, Connor,
  Bauer, and Manning}]{silveira-etal-2014-gold}
Natalia Silveira, Timothy Dozat, Marie-Catherine de~Marneffe, Samuel Bowman,
  Miriam Connor, John Bauer, and Chris Manning. 2014.
\newblock \href
  {http://www.lrec-conf.org/proceedings/lrec2014/pdf/1089_Paper.pdf} {A gold
  standard dependency corpus for {E}nglish}.
\newblock In \emph{Proceedings of the Ninth International Conference on
  Language Resources and Evaluation ({LREC}'14)}, pages 2897--2904, Reykjavik,
  Iceland. European Language Resources Association (ELRA).

\bibitem[{Sohn et~al.(2012)Sohn, Wu, and Chute}]{SohnDepParser}
Sunghwan Sohn, Stephen Wu, and Christopher Chute. 2012.
\newblock \href
  {https://doi.org/https://www.ncbi.nlm.nih.gov/pmc/articles/PMC3392064/}
  {Dependency parser-based negation detection in clinical narratives}.
\newblock \emph{AMIA Summits on Translational Science proceedings AMIA Summit
  on Translational Science}, 2012:1--8.

\bibitem[{Szarvas et~al.(2008)Szarvas, Vincze, Farkas, and
  Csirik}]{szarvas-etal-2008-bioscope}
Gy{\"o}rgy Szarvas, Veronika Vincze, Rich{\'a}rd Farkas, and J{\'a}nos Csirik.
  2008.
\newblock \href {https://www.aclweb.org/anthology/W08-0606} {The {B}io{S}cope
  corpus: annotation for negation, uncertainty and their scope in biomedical
  texts}.
\newblock In \emph{Proceedings of the Workshop on Current Trends in Biomedical
  Natural Language Processing}, pages 38--45, Columbus, Ohio. Association for
  Computational Linguistics.

\bibitem[{Taboada et~al.(2006)Taboada, Anthony, and
  Voll}]{taboada-etal-2006-methods}
Maite Taboada, Caroline Anthony, and Kimberly Voll. 2006.
\newblock \href {http://www.lrec-conf.org/proceedings/lrec2006/pdf/420_pdf.pdf}
  {Methods for creating semantic orientation dictionaries}.
\newblock In \emph{Proceedings of the Fifth International Conference on
  Language Resources and Evaluation ({LREC}{'}06)}, Genoa, Italy. European
  Language Resources Association (ELRA).

\bibitem[{Velldal et~al.(2012)Velldal, {\O}vrelid, Read, and
  Oepen}]{velldal-etal-2012-speculation}
Erik Velldal, Lilja {\O}vrelid, Jonathon Read, and Stephan Oepen. 2012.
\newblock \href {https://doi.org/10.1162/COLI_a_00126} {Speculation and
  negation: Rules, rankers, and the role of syntax}.
\newblock \emph{Computational Linguistics}, 38(2):369--410.

\bibitem[{Wiegand et~al.(2010)Wiegand, Balahur, Roth, Klakow, and
  Montoyo}]{wiegand-etal-2010-survey}
Michael Wiegand, Alexandra Balahur, Benjamin Roth, Dietrich Klakow, and
  Andr{\'e}s Montoyo. 2010.
\newblock \href {https://www.aclweb.org/anthology/W10-3111} {A survey on the
  role of negation in sentiment analysis}.
\newblock In \emph{Proceedings of the Workshop on Negation and Speculation in
  Natural Language Processing}, pages 60--68, Uppsala, Sweden. University of
  Antwerp.

\bibitem[{Wolf et~al.(2020)Wolf, Debut, Sanh, Chaumond, Delangue, Moi, Cistac,
  Rault, Louf, Funtowicz, Davison, Shleifer, von Platen, Ma, Jernite, Plu, Xu,
  Le~Scao, Gugger, Drame, Lhoest, and Rush}]{wolf-etal-2020-transformers}
Thomas Wolf, Lysandre Debut, Victor Sanh, Julien Chaumond, Clement Delangue,
  Anthony Moi, Pierric Cistac, Tim Rault, Remi Louf, Morgan Funtowicz, Joe
  Davison, Sam Shleifer, Patrick von Platen, Clara Ma, Yacine Jernite, Julien
  Plu, Canwen Xu, Teven Le~Scao, Sylvain Gugger, Mariama Drame, Quentin Lhoest,
  and Alexander Rush. 2020.
\newblock \href {https://doi.org/10.18653/v1/2020.emnlp-demos.6} {Transformers:
  State-of-the-art natural language processing}.
\newblock In \emph{Proceedings of the 2020 Conference on Empirical Methods in
  Natural Language Processing: System Demonstrations}, pages 38--45, Online.
  Association for Computational Linguistics.

\bibitem[{Zou et~al.(2015)Zou, Zhu, and Zhou}]{zou-etal-2015-negation}
Bowei Zou, Qiaoming Zhu, and Guodong Zhou. 2015.
\newblock \href {https://doi.org/10.3115/v1/P15-1064} {Negation and speculation
  identification in {C}hinese language}.
\newblock In \emph{Proceedings of the 53rd Annual Meeting of the Association
  for Computational Linguistics and the 7th International Joint Conference on
  Natural Language Processing (Volume 1: Long Papers)}, pages 656--665,
  Beijing, China. Association for Computational Linguistics.

\end{thebibliography}

\appendix

\clearpage
\section{Negation-Instance Based vs. Token-Level Weighting}
\label{sec:neg-inst-formulas}

We here explain the correspondence between NIS\textsubscript{tok} and scope token-level scoring (ST) as employed in the *SEM 2012 task in greater detail.
We use precision for our example, recall is computed analogously.

First, let us define the normalizing constant $Z$ as the sum over all predicted scope lengths.

\begin{center}
$Z = \sum_{(c_p, s_p) \in I_p} |s_p|$
\end{center}\vspace*{2mm}

For ST, precision is computed as:
\vspace*{2mm}

$\displaystyle P_{tok} = \frac{ \sum_{(c_g,s_g) \in I_g, (c_p, s_p) \in I_p, c_g = c_p} |s_g \cap s_p|}{Z}$
\vspace*{2mm}

For NIS\textsubscript{tok}, we compute precision as follows (in this formulation for simplicity disregarding the case that $s_p$ could be empty).

\noindent
$\displaystyle
P_{inst} =
\displaystyle \sum_{(c_p,s_p) \in I_p, (c_g, s_g) \in I_g, c_g = c_p} \frac{1}{|I_p|} \cdot \frac{|s_g \cap s_p|}{|s_p|}$\vspace{3mm}

If we change the weighting of the scores that each instance contributes to the sum from uniform ($\frac{1}{|I_p|}$) to a weighting scheme that weights instances by their scope length ($\frac{|s_p|}{Z}$), we arrive at the token-level metric:

\noindent
$\displaystyle
P_{tok} =
\displaystyle \sum_{(c_p,s_p) \in I_p, (c_g, s_g) \in I_g, c_g = c_p} \frac{|s_p|}{Z} \cdot \frac{|s_g \cap s_p|}{|s_p|}$\vspace{3mm}

$\displaystyle
\hspace{4mm} =
\displaystyle \sum_{(c_p,s_p) \in I_p, (c_g, s_g) \in I_g, c_g = c_p} \frac{1}{Z} \cdot |s_g \cap s_p|$
\vspace{3mm}

$\displaystyle
\hspace{4mm} =
\displaystyle \frac{1}{Z} \sum_{(c_p,s_p) \in I_p, (c_g, s_g) \in I_g, c_g = c_p} |s_g \cap s_p|$
\vspace{3mm}

$\displaystyle
\hspace{4mm} =
\displaystyle \frac{ \sum_{(c_g,s_g) \in I_g, (c_p, s_p) \in I_p, c_g = c_p} |s_g \cap s_p|}{Z}$

\vspace{0.5cm}

\noindent which corresponds to the definition of ST above.

\section{Hyperparameters}
\label{sec:hyperparameters}
This section describes the hyperparameters used in the systems implemented by us, i.e., the dependency parsers and sequence taggers used for negation resolution.

\subsection{XLM-R language model}
For the underlying XLM-R language model, we use the same set of hyperparameters in all of our experiments.
These are shown in \tref{hypers_xlmr}.

\begin{table}[h]
\centering
\footnotesize
\begin{tabular}{lr}
\toprule
Model & XLM-R-large\\
Token mask probability  & 0.15 \\
Layer dropout & 0.1 \\
Hidden dropout & 0.2 \\
Attention dropout & 0.2 \\
Output dropout & 0.5 \\
\bottomrule
\end{tabular}
\caption{Hyperparameter values for the XLM-R language model.}
\label{tab:hypers_xlmr}
\end{table}

\subsection{BiLSTM-Tagger}
\paragraph{Cue tagging.}
For the cue tagging subsystem, we simply use a linear layer with softmax on top of the XLM-R model.
The model is then trained using the hyperparameters shown in \tref{hypers_cue_tagger}.

\begin{table}[h]
\centering
\footnotesize
\begin{tabular}{lr}
\toprule
Optimizer & AdamW \\
Weight decay  &  0 \\
Batch size & 8\\
Base learning rate & $2e^{-5}$\\
LR schedule  & Noam \\
LR warmup  & 1 epoch \\
\bottomrule
\end{tabular}
\caption{Hyperparameter values for cue tagger.}
\label{tab:hypers_cue_tagger}
\end{table}

\paragraph{Scope tagging.}
For the scope tagging subsystem, we add a 1-layer BiLSTM on top of the XLM-R model and then use a linear layer with softmax to classify tokens as part of negation scopes.
In this system, we use different learning rates for the XLM-R model vs. the BiLSTM and the classifier.
We found that using low learning rates for the entire system causes it to underfit.
Furthermore, because the scope tagger is only trained on a smaller number of instances (i.e., only those that actually contain negation instances), we found it beneficial to reduce the batch size compared to the cue tagger.
Our final hyperparameters can be found in \tref{hypers_scope_tagger}.

\begin{table}[h]
\centering
\footnotesize
\begin{tabular}{lr}
\toprule
\multicolumn{2}{c}{\textbf{BiLSTM}} \\
\midrule
BiLSTM layers & 1\\
BiLSTM hidden size & 2 $\times$ 200\\
BiLSTM dropout & 0.0\\
\textit{cue}/\textit{notcue} embedding dim. & 128\\
\midrule
\multicolumn{2}{c}{\textbf{Optimization}} \\
\midrule
Optimizer & AdamW \\
Weight decay  &  0 \\
Batch size & 4\\
Base learning rate (BiLSTM) & $2e^{-4}$\\
Base learning rate (XLM-R) & $2e^{-5}$\\
LR schedule  & Noam \\
LR warmup  & 1 epoch \\
\bottomrule
\end{tabular}
\caption{Hyperparameter values for scope tagger.}
\label{tab:hypers_scope_tagger}
\end{table}

\subsection{Dependency Parser}
Our parser implementation is based on the STEPS parser by \citet{gruenewald2021applying}, which in turn is based on the unfactorized graph parsing approach by \citet{dozat-manning-2018-simpler}.
\tref{hypers_parser} gives the hyperparameters used for this system.
Like the cue tagger, we use only a single learning rate for the entire system (XLM-R model as well as classifier).

\begin{table}[h]
\centering
\footnotesize
\begin{tabular}{lr}
\toprule
\multicolumn{2}{c}{\textbf{Biaffine classifier}} \\
\midrule
Arc and label scorer dimension & 1024 \\
Dropout & 0.33 \\
\midrule
\multicolumn{2}{c}{\textbf{Optimization}} \\
\midrule
Optimizer & AdamW \\
Weight decay  &  0 \\
Batch size  & 32 \\
Base learning rate  & $4e^{-5}$ \\
LR schedule  & Noam \\
LR warmup  & 1 epoch \\
\bottomrule
\end{tabular}
\caption{Hyperparameter values for dependency parsing-based system.}
\label{tab:hypers_parser}
\end{table}

\section{Additional Tables and Figures}
\label{sec:additional_figures_tables}
\tref{full-results-appendix} contains the full set of experimental results of our study.

\fref{counts_cdneg}, \fref{counts_bioscope}, and \fref{counts_sfu} show negation instances by scope lengths for the three datasets used in our experiments.

\tref{data_negspec} gives details on the datasets annotated for negation used in our study, while \tref{data_split_stats} provides statistics for our data splits.
In addition, we will publicly release the \textbf{exact splits} (document IDs per dataset) upon publication of our paper.

\begin{table*}
\footnotesize
\centering
\begin{tabular}{lll|r|rr|rrr|rrr}
\toprule
\textbf{}&\textbf{}&\textbf{CD}&\textbf{Cues-B}&\textbf{SCM}&\textbf{SCM-B (NIS\textsubscript{ex})}&\textbf{ST}&\textbf{ST}&\textbf{ST}&\textbf{\nispar}&\textbf{\nispar}&\textbf{\nispar}\\
\textbf{System} & Train & Test & F\textsubscript{1} & F\textsubscript{1} &            F\textsubscript{1} &             P &             R & F\textsubscript{1} &             P &             R & F\textsubscript{1} \\
\midrule
\textit{Punct-BL}&\textit{--}&\textit{CD}&\textit{100.0}&\textit{17.0}&\textit{9.9}&\textit{89.8}&\textit{62.0}&\textit{73.3}&\textit{93.3}&\textit{58.8}&\textit{72.1}\\
           NegBERT &    CD &   CD &               91.1 &               78.8 &                 \textbf{70.1} &          86.1 & \textbf{87.8} &               86.9 &          83.9 & \textbf{88.7} &               86.3 \\
         BiLSTM-Tagger &    CD &   CD &               92.3 &      \textbf{79.6} &                          70.0 & \textbf{90.8} &          85.4 &      \textbf{88.0} & \textbf{90.3} &          86.3 &      \textbf{88.2} \\
      DP-direct &    CD &   CD &               92.8 &               73.2 &                          61.4 &          85.4 &          87.4 &               86.4 &          84.6 &          87.7 &               86.1 \\
      DP-nested &    CD &   CD &               93.3 &               72.8 &                          60.6 &          86.6 &          83.0 &               84.8 &          85.6 &          85.4 &               85.5 \\
 DP-direct-synt &    CD &   CD &               92.8 &               79.1 &                          69.2 &          87.4 &          86.5 &               86.9 &          86.2 &          88.3 &               87.2 \\
 DP-nested-synt &    CD &   CD &      \textbf{93.4} &               79.2 &                          69.2 &          88.0 &          84.6 &               86.2 &          87.2 &          86.6 &               86.9 \\
\midrule
           NegBERT &   Bio &   CD &               65.2 &               13.0 &                           8.9 &          75.6 &          43.7 &               55.2 &          71.1 &          40.0 &               51.1 \\
         BiLSTM-Tagger &   Bio &   CD &               61.6 &               14.7 &                          10.1 &          72.1 &          33.8 &               46.0 &          70.5 &          35.0 &               46.8 \\
      DP-direct &   Bio &   CD &               65.2 &               13.1 &                           8.9 &          74.1 &          36.2 &               48.5 &          74.3 &          35.1 &               47.6 \\
      DP-nested &   Bio &   CD &               63.2 &               13.6 &                           9.3 &          79.2 &          32.9 &               46.5 &          77.2 &          31.9 &               45.1 \\
 DP-direct-synt &   Bio &   CD &               65.6 &               13.8 &                           9.1 &          76.5 &          37.8 &               50.5 &          76.4 &          34.9 &               47.8 \\
 DP-nested-synt &   Bio &   CD &               64.8 &               13.7 &                           9.2 &          80.1 &          38.4 &               51.8 &          77.7 &          34.7 &               47.9 \\
\midrule
           NegBERT &   SFU &   CD &               69.3 &               15.2 &                           9.7 &          74.4 &          46.1 &               56.7 &          70.2 &          43.6 &               53.6 \\
         BiLSTM-Tagger &   SFU &   CD &               68.8 &               15.7 &                          10.1 &          78.3 &          46.0 &               58.0 &          74.0 &          42.3 &               53.8 \\
      DP-direct &   SFU &   CD &               69.2 &               14.2 &                           9.1 &          78.0 &          42.8 &               55.3 &          74.9 &          40.0 &               52.1 \\
      DP-nested &   SFU &   CD &               68.2 &               14.0 &                           9.0 &          77.7 &          42.4 &               54.8 &          74.7 &          38.6 &               50.9 \\
 DP-direct-synt &   SFU &   CD &               67.5 &               15.0 &                           9.7 &          77.7 &          44.3 &               56.5 &          73.1 &          39.7 &               51.4 \\
 DP-nested-synt &   SFU &   CD &               68.1 &               15.0 &                           9.6 &          77.0 &          43.1 &               55.2 &          72.9 &          39.9 &               51.6 \\
\midrule
\textit{Punct-BL}&\textit{--}&\textit{Bio}&\textit{100.0}&\textit{59.4}&\textit{43.2}&\textit{67.3}&\textit{64.2}&\textit{65.7}&\textit{77.0}&\textit{69.7}&\textit{73.5}\\
           NegBERT &   Bio &  Bio &               90.5 &               85.6 &                          79.5 &          81.0 &          88.3 &               84.5 &          83.3 &          91.5 &               87.2 \\
         BiLSTM-Tagger &   Bio &  Bio &               94.3 &      \textbf{89.6} &                 \textbf{83.0} &          88.9 &          88.5 &               88.6 &          91.7 &          91.7 &               91.7 \\
      DP-direct &   Bio &  Bio &      \textbf{95.4} &               83.8 &                          74.3 &          85.6 &          89.2 &               87.3 &          88.7 &          92.4 &               90.5 \\
      DP-nested &   Bio &  Bio &               95.3 &               87.4 &                          79.8 &          89.1 &          87.9 &               88.5 &          91.2 &          91.2 &               91.2 \\
 DP-direct-synt &   Bio &  Bio &               95.1 &               88.2 &                          81.0 &          89.4 &          88.9 &               89.1 &          91.7 &          91.7 &               91.7 \\
 DP-nested-synt &   Bio &  Bio &      \textbf{95.4} &               89.1 &                          82.7 & \textbf{91.0} & \textbf{89.8} &      \textbf{90.4} & \textbf{92.7} & \textbf{92.1} &      \textbf{92.4} \\
\midrule
           NegBERT &    CD &  Bio &               60.6 &               24.5 &                          19.0 &          36.4 &          77.3 &               49.3 &          32.6 &          80.5 &               46.2 \\
         BiLSTM-Tagger &    CD &  Bio &               64.3 &               24.6 &                          17.7 &          45.3 &          76.1 &               56.7 &          41.2 &          80.1 &               54.3 \\
      DP-direct &    CD &  Bio &               64.4 &               17.5 &                          12.5 &          38.0 &          76.0 &               50.7 &          32.1 &          79.0 &               45.7 \\
      DP-nested &    CD &  Bio &               65.8 &               18.5 &                          13.1 &          40.0 &          78.7 &               53.0 &          34.9 &          78.9 &               48.4 \\
 DP-direct-synt &    CD &  Bio &               64.7 &               22.2 &                          16.3 &          40.3 &          78.0 &               53.1 &          34.8 &          80.3 &               48.5 \\
 DP-nested-synt &    CD &  Bio &               65.3 &               25.2 &                          18.7 &          42.2 &          77.6 &               54.7 &          37.2 &          80.6 &               50.9 \\
\midrule
           NegBERT &   SFU &  Bio &               78.7 &               64.7 &                          55.5 &          73.9 &          55.4 &               63.2 &          76.8 &          66.1 &               71.0 \\
         BiLSTM-Tagger &   SFU &  Bio &               81.0 &               64.3 &                          53.4 &          74.8 &          58.8 &               65.9 &          77.8 &          68.4 &               72.8 \\
      DP-direct &   SFU &  Bio &               81.6 &               59.6 &                          48.3 &          75.4 &          55.9 &               64.2 &          79.2 &          66.0 &               72.0 \\
      DP-nested &   SFU &  Bio &               82.1 &               61.1 &                          49.9 &          77.1 &          57.4 &               65.8 &          80.0 &          66.8 &               72.8 \\
 DP-direct-synt &   SFU &  Bio &               81.6 &               66.8 &                          57.1 &          77.3 &          59.2 &               67.0 &          80.0 &          68.4 &               73.7 \\
 DP-nested-synt &   SFU &  Bio &               82.0 &               67.0 &                          57.2 &          77.8 &          59.2 &               67.2 &          80.1 &          68.8 &               74.0 \\
\midrule
\textit{Punct-BL}&\textit{--}&\textit{SFU}&\textit{100.0}&\textit{56.9}&\textit{45.9}&\textit{55.0}&\textit{84.0}&\textit{66.5}&\textit{75.5}&\textit{87.4}&\textit{81.0}\\
           NegBERT &   SFU &  SFU &               81.9 &               74.2 &                          69.5 &          63.6 &          81.0 &               71.0 &          67.9 &          86.3 &               75.9 \\
         BiLSTM-Tagger &   SFU &  SFU &      \textbf{86.7} &      \textbf{78.0} &                 \textbf{73.2} & \textbf{71.1} & \textbf{86.0} &      \textbf{77.9} & \textbf{74.9} & \textbf{89.8} &      \textbf{81.7} \\
      DP-direct &   SFU &  SFU &               86.1 &               73.5 &                          66.9 &          71.2 &          79.5 &               75.1 &          73.2 &          86.1 &               79.1 \\
      DP-nested &   SFU &  SFU &               85.8 &               73.0 &                          66.1 &          70.7 &          80.1 &               75.0 &          70.8 &          80.2 &               75.2 \\
 DP-direct-synt &   SFU &  SFU &               86.6 &               74.6 &                          68.5 &          71.8 &          80.8 &               76.0 &          73.4 &          87.2 &               79.6 \\
 DP-nested-synt &   SFU &  SFU &               86.5 &               76.6 &                          71.1 &          72.2 &          82.7 &               77.1 &          73.7 &          88.9 &               80.5 \\
\midrule
           NegBERT &    CD &  SFU &               51.3 &               14.4 &                          10.0 &          27.2 &          58.8 &               37.2 &          26.6 &          63.6 &               37.5 \\
         BiLSTM-Tagger &    CD &  SFU &               50.2 &               13.4 &                           9.0 &          28.9 &          53.6 &               37.5 &          28.0 &          58.0 &               37.8 \\
      DP-direct &    CD &  SFU &               50.6 &               10.3 &                           6.9 &          25.9 &          53.5 &               34.9 &          26.4 &          56.9 &               36.1 \\
      DP-nested &    CD &  SFU &               50.6 &               11.6 &                           7.8 &          26.8 &          53.1 &               35.6 &          27.0 &          56.4 &               36.5 \\
 DP-direct-synt &    CD &  SFU &               50.6 &               12.3 &                           8.3 &          27.2 &          55.1 &               36.4 &          26.7 &          58.8 &               36.8 \\
 DP-nested-synt &    CD &  SFU &               50.3 &               13.2 &                           9.0 &          27.2 &          54.1 &               36.2 &          26.6 &          58.0 &               36.5 \\
\midrule
           NegBERT &   Bio &  SFU &               63.4 &               54.0 &                          48.8 &          60.1 &          49.7 &               53.7 &          63.6 &          55.4 &               58.6 \\
         BiLSTM-Tagger &   Bio &  SFU &               64.5 &               56.6 &                          53.1 &          67.9 &          49.1 &               57.0 &          70.2 &          54.5 &               61.4 \\
      DP-direct &   Bio &  SFU &               66.6 &               51.9 &                          45.9 &          63.2 &          49.2 &               55.2 &          65.1 &          54.7 &               59.3 \\
      DP-nested &   Bio &  SFU &               65.1 &               51.2 &                          45.9 &          64.0 &          44.7 &               52.6 &          66.9 &          51.2 &               58.0 \\
 DP-direct-synt &   Bio &  SFU &               56.1 &               48.6 &                          39.4 &          46.6 &          47.7 &               47.0 &          48.6 &          53.5 &               50.8 \\
 DP-nested-synt &   Bio &  SFU &               56.9 &               48.6 &                          40.1 &          48.2 &          47.0 &               47.5 &          50.7 &          52.5 &               51.5 \\
\bottomrule
\end{tabular}
\caption{Complete experimental results: For cues, we report the *SEM 2012 B version for exact cue matching \textbf{(Cues-B)}. \textbf{SCM} refers to the standard version of Scope Level F1 of *SEM 2012, \textbf{SCM-B} to their B-version, which also corresponds to our \textbf{NIS\textsubscript{ex}}. The punctuation baseline \textit{Punct-BL} uses gold cues and predicts scopes as starting from a cue to the next punctuation token.}
\label{tab:full-results-appendix}
\end{table*}

\begin{table*}[]
\centering
\footnotesize
\begin{tabular}{l|C{85pt}|C{85pt}|C{87pt}}
	\toprule
	\textbf{Dataset} & \textbf{\textit{ConanDoyle-neg}} & \textbf{\textit{BioScope Abstracts}} & \textbf{\textit{SFU Review}} \\
	\midrule
	\textbf{Source} & \citet{morante-daelemans-2012-conandoyle} & \citet{szarvas-etal-2008-bioscope} & \citet{konstantinova-etal-2012-review} \\
	\midrule
	\textbf{Domain} & fiction writing & biomedical & review \\
	\textbf{Sentence $\#$} & 5,520 & 11,871 & 17,263 \\
	\textbf{Negation sentence $\#$} & 1,227 & 1,597 & 3,117 \\
	\textbf{Negation instance $\#$} & 1,421 (original) / 1,432 (ours) & 1,719 & 3,518 \\ 
	\textbf{Annotated for speculation} & no & yes & yes \\
	\textbf{Cue is a part of the scope} & no & yes & no \\
	\textbf{Includes discontinuous scopes} & yes & no & yes \\
	\textbf{Includes events} & yes & no & no \\
	\textbf{Annotates negation affixes} & yes & rarely, with the whole word as a cue & yes, but with the whole word as a cue \\
	\textbf{Tokenized} & yes & no & yes \\
	\textbf{File format} & CoNLL & XML & XML \\
	\bottomrule
\end{tabular}
\caption{Overview of datasets annotated for negation used in our study.}
\label{tab:data_negspec}
\end{table*}

\begin{table*}[t]
\centering
{\footnotesize
\begin{tabular}{lll|rrr|r}
\toprule
\multicolumn{3}{l|}{\textbf{Dataset}} & \textbf{Train} & \textbf{Dev} & \textbf{Test} & \textbf{Total} \\
\midrule
\multicolumn{2}{l}{\multirow{5}{*}{\textit{ConanDoyle-neg}}} & sentence $\#$ & 3,644 & 787 & 1,089 & 5,520 \\
&& sentence \% & 66\% & 14.3\% & 19.7\% & 100 \% \\
\cline{3-7}
&& negation instance $\#$ & 984 & 173 & 264 & 1,421 \\
&& negation sentence $\#$ & 848 & 144 & 235 & 1,227 \\
&& negation sentence \% & 23.3\% & 18.3\% & 21.6\% & 22.2 \% \\
\cline{3-7}
\multicolumn{2}{l}{(reannotated)} & negation instance $\#$ & 987 & 176 & 269 & 1,432 \\
\midrule
\multirow{5}{*}{\textit{BioScope Abstracts}} &  & sentence $\#$ & 9,500 & 1,185 & 1,186 & 11,871 \\
&& sentence \% & 80\% & 10\% & 10\% & 100\% \\
\cline{3-7}
&& negation instance $\#$ & 1,396 & 156 & 167 & 1,719 \\
&& negation sentence $\#$ & 1,297 & 148 & 152 & 1,597 \\
&& negation sentence \% & 13.7\% & 12.5\% & 12.8\% & 13.5\% \\
\midrule
\multicolumn{2}{l}{\multirow{5}{*}{\textit{SFU Review}}} & sentence $\#$ & 13,614 & 1,817 & 1,800 & 17,231 \\
&& sentence \% & 79\% & 11\% & 10\% & 100\% \\
\cline{3-7}
&& negation instance $\#$ & 2,835 & 365 & 309 & 3,509 \\
&& negation sentence $\#$ & 2,503 & 328 & 276 & 3,107 \\
&& negation sentence \% & 18.4\% & 18.1\% & 15.3\% & 18\% \\
\bottomrule
\end{tabular}
}
\caption{Dataset splits}
\label{tab:data_split_stats}
\end{table*}

\begin{figure}
    \centering
    \hspace{-0.35cm}
    \includegraphics[scale=0.49]{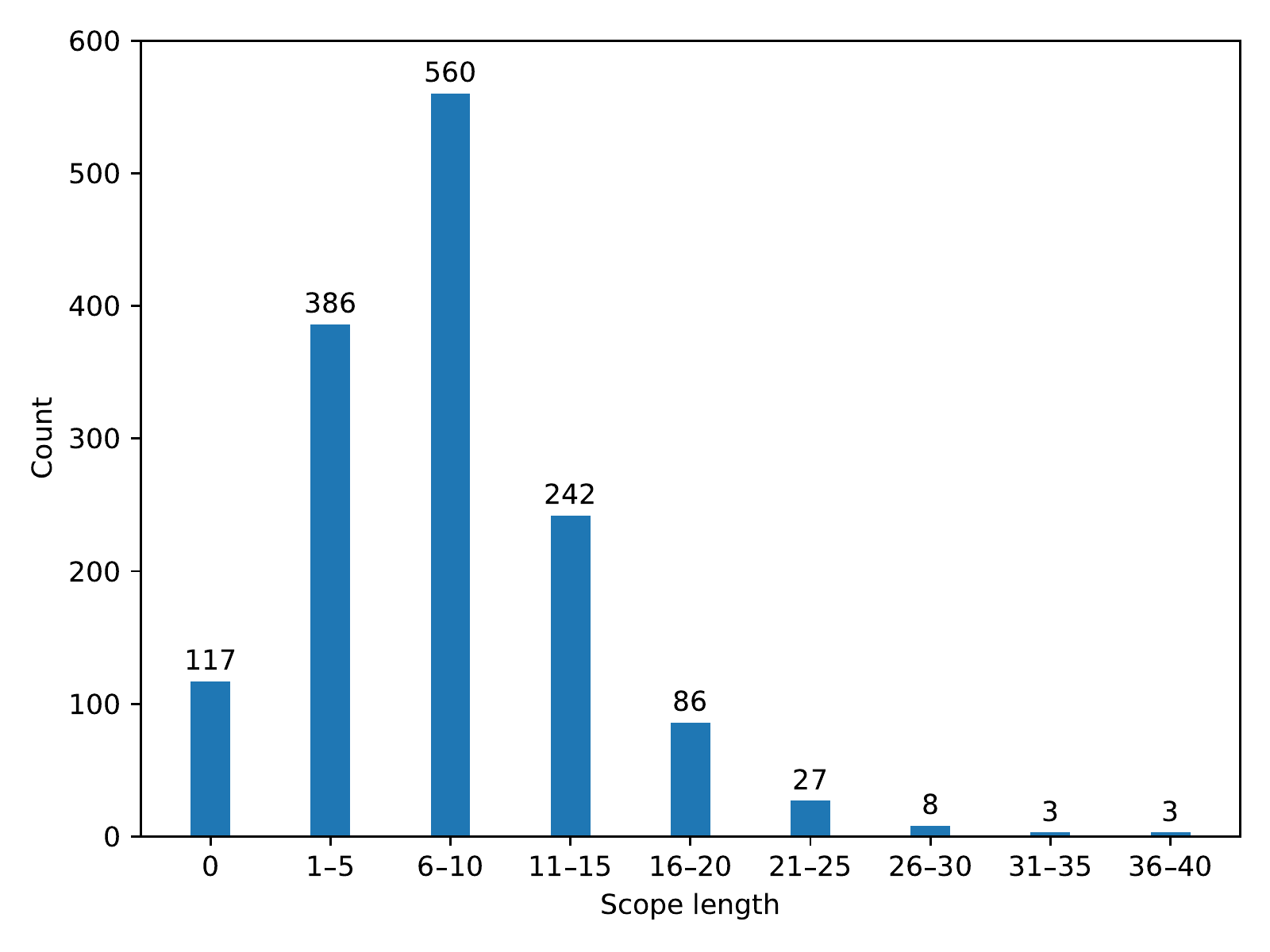}
    \caption{Negation instances by scope length for CD-neg.}
    \label{fig:counts_cdneg}
\end{figure}

\begin{figure}[t]
    \centering
    \hspace{-0.35cm}
    \includegraphics[scale=0.49]{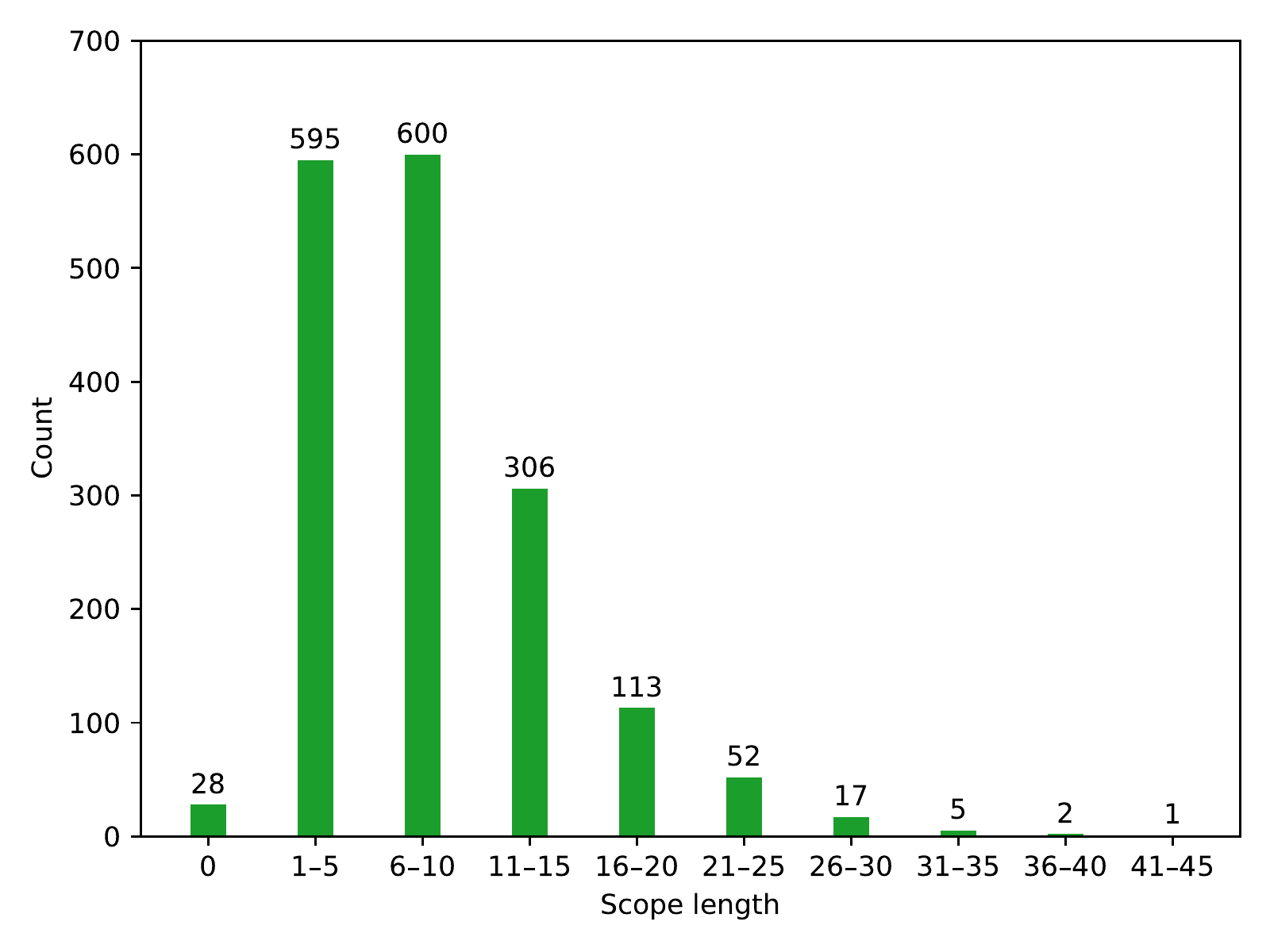}
    \caption{Negation instances by scope length for BioScope abstracts.}
    \label{fig:counts_bioscope}
\end{figure}

\begin{figure}[t]
    \centering
    \hspace{-0.35cm}
    \includegraphics[scale=0.49]{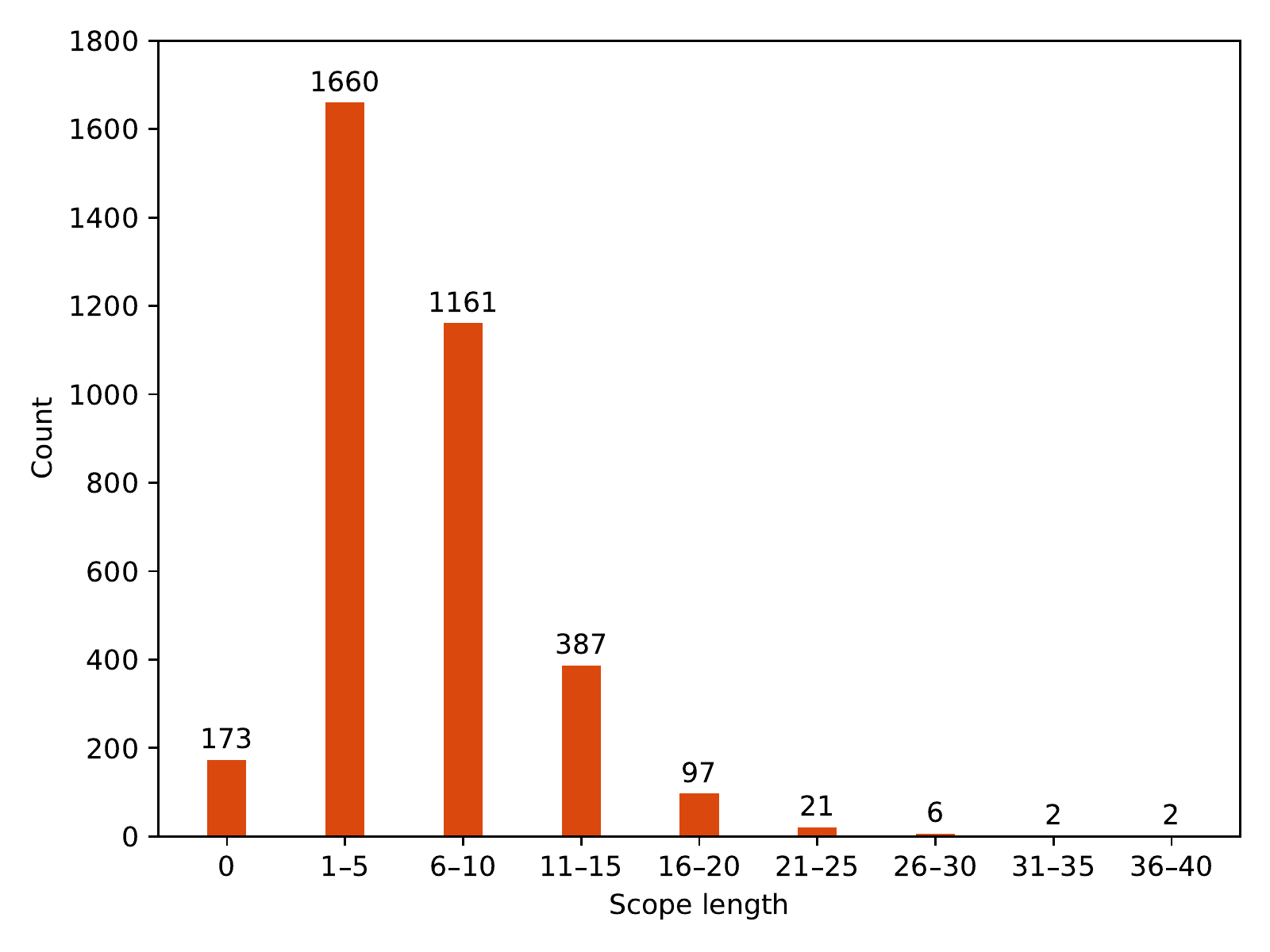}
    \caption{Negation instances by scope length for SFU Review.}
    \label{fig:counts_sfu}
\end{figure}

\end{document}